\documentclass[10pt,twocolumn,letterpaper]{article}

\pdfoutput=1
\usepackage[pagenumbers]{cvpr} % To force page numbers, e.g. for an arXiv version
\usepackage{amsmath}
\usepackage{amssymb}
\usepackage{algorithm}
\usepackage{algpseudocode}
\usepackage{multirow}
\usepackage{subcaption}
\usepackage[page]{appendix} % print appendices title
\usepackage
{soul}
\usepackage[table]{xcolor}
\usepackage{rotating}
\makeatletter
\robustify\@latex@warning@no@line
\makeatother
\usepackage{authblk}

\definecolor{cvprblue}{rgb}{0.21,0.49,0.74}
\usepackage[pagebackref,breaklinks,colorlinks,citecolor=cvprblue]{hyperref}

\def\paperID{12449} % *** Enter the Paper ID here
\def\confName{CVPR}
\def\confYear{2024}

\title{IL-NeRF: Incremental Learning for Neural Radiance Fields with Camera Pose Alignment}

\author[1]{Letian Zhang}
\author[2]{Ming Li}
\author[2]{Chen Chen}
\author[3]{Jie Xu}
\vspace{-50 pt}
\affil[1]{Department of Computer Science, Middle Tennessee State University}
\affil[2]{Center for Research in Computer Vision, University of Central Florida}
\affil[3]{Department of Electrical and Computer Engineering,
University of Miami}

\begin{document}
\setlength{\abovedisplayskip}{2pt}
\setlength{\belowdisplayskip}{2pt}

\maketitle
\begin{abstract}
% Neural radiance fields (NeRF) is a promising approach for generating photorealistic images and representing complex scenes. However, when processing data sequentially, it can suffer from catastrophic forgetting, where previous data is easily forgotten after training with new data. While existing works propose effective incremental learning methods through knowledge distillation, these approaches are founded on the assumption that continuous data chunks comprise not only 2D images but also corresponding camera pose parameters, pre-estimated from the complete dataset. This assumption poses a paradox as the required camera pose must be estimated from the entire dataset, yet the data is intended to arrive sequentially, relatively independently, with future data chunks remaining unknown and inaccessible while previous chunks are discarded. In contrast, we address a more practical scenario where camera poses are unknown. In this work, we present a novel framework for incremental NeRF training, named IL-NeRF.
% The core concept of IL-NeRF involves appropriately selecting a set of past camera poses as references to initialize and align the camera poses of newly arriving image data. This is followed by a joint optimization of both camera poses and replay-based NeRF distillation.
% Our experiments on real-world indoor and outdoor scenes demonstrate that IL-NeRF handles incremental learning for NeRF and outperforms baselines by up to $54.04\%$ in rendering quality.

Neural radiance fields (NeRF) is a promising approach for generating photorealistic images and representing complex scenes. However, when processing data sequentially, it can suffer from catastrophic forgetting, where previous data is easily forgotten after training with new data. Existing incremental learning methods using knowledge distillation assume that continuous data chunks contain both 2D images and corresponding camera pose parameters, pre-estimated from the complete dataset. This poses a paradox as the necessary camera pose must be estimated from the entire dataset, even though the data arrives sequentially and future chunks are inaccessible. In contrast, we focus on a practical scenario where camera poses are unknown. We propose IL-NeRF, a novel framework for incremental NeRF training, to address this challenge. 
IL-NeRF's key idea lies in selecting a set of past camera poses as references to initialize and align the camera poses of incoming image data. This is followed by a joint optimization of camera poses and replay-based NeRF distillation. Our experiments on real-world indoor and outdoor scenes show that IL-NeRF handles incremental NeRF training and outperforms the baselines by up to $54.04\%$ in rendering quality. The project page is \url{https://ilnerf.github.io/}.

\end{abstract}
\vspace{-0.5cm}
\section{Introduction}
\label{sec:intro}
% In the fields of AR/VR \cite{gafni2021dynamic}, embedded artificial intelligence \cite{ma2019accurate}, and robotics \cite{durrant2006simultaneous}, it is essential to be able to reconstruct real-world objects using RGB cameras. 
Neural Radiance Fields (NeRF) \cite{mildenhall2021nerf} has recently shown great promise in producing photorealistic images from sparse image sets by encoding a 3D scene with a neural network that maps the location of 3D points to color and volume density. 
However, to achieve this level of performance, NeRF typically requires access to all training data at once in order to estimate the camera pose based on the entire dataset \cite{chung2022meil,mildenhall2021nerf}. In practical applications such as automotive and remote sensing, data is acquired sequentially, necessitating an immediately available updated 3D scene representation.  Moreover, scenarios arise where a user acquires scans of a scene to train NeRF, only to find that the training yielded less effective results than expected. Consequently, the user rescans the scene with new data to enhance the fidelity of the images rendered by NeRF. In these instances, the scene representation must be trained in an incremental training environment, where the model can only access a limited number of views at each training stage, while still undertaking the task of reconstructing the entire scene.
\begin{figure}[tt]
	\centering
	\includegraphics[width=\linewidth]{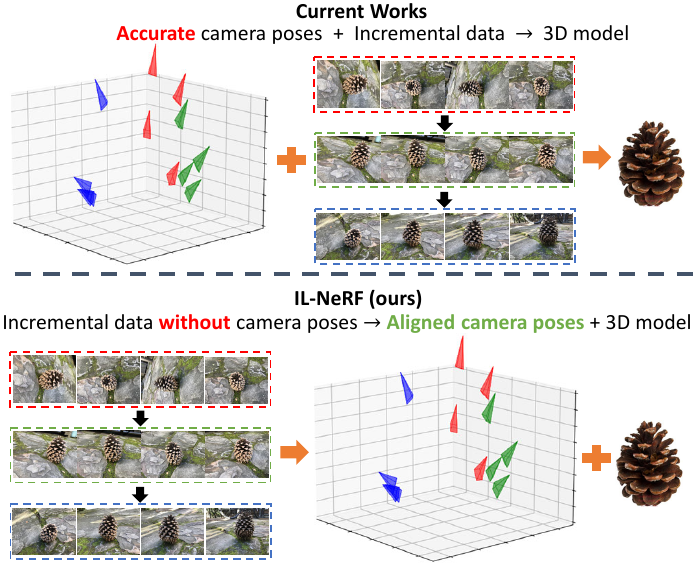}
 \vspace{-0.7cm}
	\caption{Current works require accurate camera poses pre-estimated from the whole image data. Our IL-NeRF can incrementally learn the 3D scene without camera poses. The results of IL-NeRF consist of both aligned camera poses and NeRF model.}
	\label{fig:compareILNeRFandCurrent}
	\vspace{-14 pt}
\end{figure}

In the context of incremental NeRF training, existing works \cite{guo2022incremental,chung2022meil,po2023instant,cai2023clnerf} generally operate under the assumption that data is segmented into multiple sequential chunks, with access limited to the current chunk while previously processed data is discarded. This assumption to incremental learning presents a notable challenge for NeRF, which requires updating its knowledge without erasing prior learned information, a phenomenon known as catastrophic forgetting \cite{zhang2019your}. To address this challenge, prior works \cite{guo2022incremental,chung2022meil,po2023instant,cai2023clnerf} have investigated the implementation of incremental learning strategies for NeRF training, incorporating a knowledge distillation technique \cite{zhang2019your} to mitigate catastrophic forgetting. Specifically, before training the NeRF model with the current data chunk, pseudo-RGB values for the scene are generated using the previously trained NeRF model. These RGB values are subsequently utilized to train the NeRF model with the current data chunk, and the process is iteratively repeated, with the prior NeRF model acting as the supervisory teacher for subsequent data chunks. This enables the NeRF model to learn from new data while retaining knowledge from previously discarded data.

\textbf{Motivation.} While existing works propose effective incremental learning methods through knowledge distillation, these approaches are founded on the assumption that continuous data chunks comprise not only 2D images \textit{but also corresponding camera pose parameters, pre-estimated from the complete dataset}. This assumption poses a paradox as the required camera pose must be estimated from the entire dataset, yet the data is intended to arrive sequentially, relatively independently, with future data chunks remaining unknown and inaccessible while previous chunks are discarded.
\ul{In contrast, as shown in Figure {\ref{fig:compareILNeRFandCurrent}}, our work addresses a more practical scenario where pre-estimated camera poses are unavailable for each training data chunk, necessitating the consideration of camera pose estimation and the alignment of its coordinate system}.

\textbf{Challenge.} Since the previous training data have been discarded, the incoming training data cannot simply be used directly for camera pose estimation because the isolated estimated camera pose will not be in the same coordinate system as the previous camera pose, which will lead to NeRF training misalignment and failure to render the 3D scene. Therefore, accurately estimating the camera poses of the sequential coming data within the same coordinate system in incremental NeRF training becomes a crucial issue that needs to be addressed.

\textbf{Contribution.} 
% In this work, we present a novel framework for incremental NeRF training, named IL-NeRF, which can effectively tackle both the challenge of catastrophic forgetting and the necessity to estimate and align camera poses within a consistent coordinate system.
To deal with the above challenge, we present a novel framework for incremental NeRF training, named IL-NeRF, which can 
incrementally estimate the incoming data's camera poses and effectively tackle the issue of catastrophic forgetting.
% (1) We propose an incremental camera pose alignment module that selects optimal previous camera poses to render prior training images from NeRF and assist in joint camera pose estimation. This ensures the consistent alignment of new and existing camera poses.
(1) We propose an incremental camera pose alignment module that selects a suitable set of camera poses from previous estimates. These chosen poses help render prior training images from NeRF and aid in the joint camera pose estimation process. They also act as a reference coordinate system, facilitating the alignment of newly arrived and previous camera poses.
% (2) We transform the selection of camera poses into a reward-collection optimization problem based on a graph. We then introduce a practical greedy algorithm to effectively solve this optimization issue.
(2) To ensure the appropriate selection of camera poses from the previously estimated ones, we transform this selection task into a graph-based reward-collection optimization problem. We then introduce a practical greedy algorithm to effectively solve this optimization problem.
(3) To align camera pose coordinates, we use selected camera poses as a reference to derive a transfer matrix, transforming the current camera poses into the previous coordinate system.
% (3) To align the camera pose coordinate system, we derive a transformation matrix using the selected camera poses as a reference, enabling the transformation of the current data's camera poses into the coordinate system of the previous data's camera poses.
(4) We utilize a joint optimization method for camera poses and replay-based NeRF distillation, mitigating catastrophic forgetting and refining the accuracy of the camera poses.
% (4) To enhance the precision of camera pose estimation, we employ a joint optimization training method for both camera poses and replay-based NeRF distillation. This approach addresses the catastrophic forgetting challenge while also refining the accuracy of the camera poses.

The experimental results on three diverse datasets show that our proposed framework can improve PSNR, SSIM, and LPIPS by up to $54.04\%$ compared to the original NeRF, significantly mitigating the negative impact of catastrophic forgetting in NeRF's incremental learning process. Moreover, our framework can effectively estimate and align the camera pose parameters in a consistent coordinate system.

\vspace{-5 pt}

\section{Related Works}
\label{sec:related_work}

\textbf{NeRF.}
NeRF \cite{mildenhall2021nerf} employs volume rendering to depict a continuous scene and achieve high-quality view synthesis. Several subsequent works have been introduced based on the success of NeRF, aiming to improve view synthesis efficiency and quality, including faster training and rendering \cite{muller2022instant,chen2022tensorf,fridovich2022plenoxels}, deformable or dynamic scene synthesis \cite{park2021nerfies,pumarola2021d,rebain2021derf}, editable view synthesis \cite{niemeyer2021giraffe,xiang2021neutex}, light changes \cite{martin2021nerf,mildenhall2022nerf}, surface enhancements \cite{oechsle2021unisurf,wang2021neus,yariv2021volume}, depth priors \cite{deng2022depth,roessle2022dense,wei2021nerfingmvs}, etc. However, most of these methods presume access to all training data and require pre-estimated camera pose parameters from the entire dataset. In this study, we tackle a more practical scenario where NeRF learns the scene incrementally with a sequential data stream, without pre-estimated camera pose parameters. \\
\textbf{Incremental NeRF Learning.} 
Incremental learning, constrained by limited data availability during each training iteration, often causes catastrophic forgetting \cite{french1999catastrophic}. Methods to mitigate this issue include parameter isolation \cite{hung2019compacting,mallya2018piggyback,mallya2018packnet}, replay \cite{lopez2017gradient,rolnick2019experience,shin2017continual}, and regularization \cite{aljundi2017expert,xu2018reinforced,aljundi2018memory}. 
There is limited research on combining NeRF with incremental learning. Existing studies \cite{guo2022incremental,chung2022meil,po2023instant,cai2023clnerf} use knowledge distillation \cite{zhang2019your} to mitigate catastrophic forgetting by accessing past training data, specifically RGB values, from a pre-trained network. The retrieved data is merged with new incoming data to train NeRF.
In \cite{guo2022incremental}, a regularization-based filter is used to select relevant information from randomly sampled camera views. In \cite{chung2022meil}, a small network is trained to generate previously seen rays that are directed toward the scene. The authors in \cite{po2023instant} employ the same replay-based method in \cite{chung2022meil} but they substitute NeRF with Instant-NGP \cite{muller2022instant} to expedite the training process. In \cite{cai2023clnerf}, a prioritized replay buffer is introduced to keep the images and their camera poses with the lowest historical rendering qualities for continual learning. However, these methods require pre-estimated camera poses from the entire training data for each coming data chunk. In this work, we consider a more realistic scenario where pre-estimated camera poses are unavailable for each training data chunk, thus requiring incremental camera pose alignment. \\
\textbf{NeRF With Pose Refinement.}
Pose refinement is widely used in NeRF training. iNeRF \cite{yen2021inerf} refines camera poses for novel view images using a reconstructed NeRF model. NeRFmm \cite{wang2021nerf} jointly optimizes both camera intrinsic and extrinsic parameters during NeRF training, and BARF \cite{lin2021barf} proposes a coarse-to-fine positional encoding strategy for camera poses and NeRF joint optimization. SC-NeRF \cite{jeong2021self} further considers camera distortion refinement and employs a geometric loss to regularize rays. In this paper, IL-NeRF uses the pose refinement in SC-NeRF \cite{jeong2021self}. Note that IL-NeRF is not limited to only the pose refinement in SC-NeRF \cite{jeong2021self}; any other well-designed pose refinement methods can also be transferred to IL-NeRF.

% while GNeRF \cite{meng2021gnerf} combines GANs with NeRF to refine camera poses but requires a known sampling distribution for poses.

\section{Incremental NeRF Training Preliminary}
\label{sec:preliminary}

\textbf{NeRF.}
NeRF aims to learn a 3D scene with a simple neural network, e.g., MLPs, that takes 3D location $\textbf{x}$ and view direction $\textbf{r}_d$ as input and produces RGB color $\textbf{c}$ and volume density $\mathbf{\sigma}$ as output. For each ray $\textbf{r}= (\textbf{r}_o, \textbf{r}_d)$ emitted from the camera origin $\textbf{r}_o$ in direction $\textbf{r}_d$, NeRF samples $M$ 3D points along the ray $\textbf{x}_i = \textbf{r}_o + z_i\textbf{r}_d$, where $z_i$ is the distance from a camera to the sample point $\textbf{x}_i$. Thus the pixel color $C$ can be integrated by the volumetric rendering as follows: 
\begin{align}\label{eq:volumeRendering}
    C(\textbf{r}) = \sum_{i=1}^M \alpha_i (1-\delta_i)\textbf{c}(\textbf{x}_i)
\end{align} 
where $\delta_i = \exp(-(z_i - z_{i-1})\mathbf{\sigma}(\textbf{x}_i)$ represents the transmittance of the ray segment between sample points $\textbf{x}_{i-1}$ and $\textbf{x}_i$, and $\alpha_i = \prod_{j=1}^{i-1}\delta_j$ is the ray attenuation from the origin $\textbf{r}_o$ to the sample point $\textbf{x}_i$.
Since the whole pipeline is differentiable, NeRF can be trained by minimizing the photometric error between the rendering views and ground truth views.
\begin{align}\label{eq:nerfLossFunc}
    & \mathcal{L} = \sum_{\textbf{r} \in R} \parallel  C^*(\textbf{r}) - C(\textbf{r}) \parallel ^2_2 \\
    & {\Theta}^* = \arg \min_{\Theta} \mathcal{L}(C \mid C^*, \mathcal{P})
\end{align}
where $R$ represents a group of rays from one or more camera views, which is obtained from the camera pose parameters $\mathcal{P}$. $C^*$ is the ground truth of pixel color. $\Theta$ denotes the parameters of the network. For more details of NeRF, we refer the readers to \cite{mildenhall2021nerf}.

\textbf{Incremental NeRF Training.}
To achieve impressive performance, NeRF assumes access to all training data that covers all views of a scene at once. However, the entire training data might not be available simultaneously in practical applications due to physical or hardware limitations, e.g., edge devices with a limited amount of memory and limited data storage. As a result, data needs to be processed sequentially. In other words, NeRF will incrementally learn the scene with new training data without revisiting the old ones.
Concretely, we consider a time-slotted system wherein each time slot $t \in\{0, \cdots, T-1\}$, a chunk of image data $G^t$ consists of $N$ number of images $I^t$, i.e., $G^t = \{I_n^t, n\in\{0, \cdots N-1\}\}$. Generally, only the latest image data chunk $G^t$ is available while previous image data chunks $G^{0:t-1}$ are inaccessible. The objective of incremental NeRF is to minimize reconstruction loss across all provided
chunks of image data in $\{G_0, \dots , G_{T-1}\}$.
 Note that unlike the existing works \cite{chung2022meil,guo2022incremental,po2023instant,cai2023clnerf}, in our work, each incoming data contains only the images, without corresponding camera poses. \textit{Thus, in this paper, the main aim of incremental learning for NeRF is to enable the neural network to learn and adapt continually by ensuring that the camera poses from new image data are estimated and aligned in a consistent coordinate system while preventing catastrophic forgetting across all previously seen image data.}

\begin{figure*}[tt]
	\centering
	\includegraphics[width=0.85\linewidth]{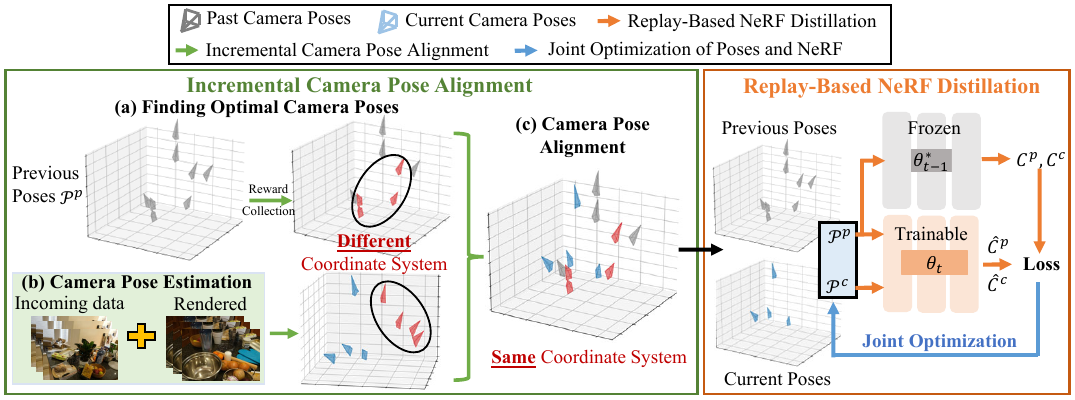}
 \vspace{-6pt}
	\caption{Overview of IL-NeRF pipeline. Firstly, the network $\Theta_{t-1}^*$ from the previous NeRF are frozen. Then, incremental camera pose alignment is employed to estimate the current camera poses $\mathcal{P}^c$ through (a) Finding optimal camera poses from the previous camera poses; (b) Estimating the camera poses for the incoming image data and the rendered images from the selected camera poses; (c) Aligning the current camera poses into the previous camera coordinate system. Finally, the network $\Theta_{t}$, the current estimated poses $\mathcal{P}^c$, and previous poses $\mathcal{P}^p$ are jointly trained on both the current image data rays $C^c$ and the distilled past rays $C^p$ simultaneously.
    % The IL-NeRF pipeline involves a multi-step process. Firstly, the network parameters $\Theta_{t-1}^*$ from the previous NeRF are frozen for replay-based NeRF distillation while training the current data chunk $G^t$. Incremental camera pose alignment is employed to estimate the current camera poses $\mathcal{P}^c$ using both the current image data and the previous training images rendered by the frozen network. The network $\Theta_{t}$, the current estimated poses $\mathcal{P}^c$, and previous poses $\mathcal{P}^p$ are then jointly trained on both the current image data rays $C^c$ and the distilled past rays $C^p$ simultaneously.
 }
	\label{fig:ilNerfPipeline}
	\vspace{-14 pt}
\end{figure*}

\section{IL-NeRF}
\label{sec:ilNerf}
In this section, we introduce our proposed framework, IL-NeRF, which prevents the catastrophic forgetting problem by replay-based knowledge distillation retrieved from NeRF itself (Section \ref{sec:knowledgeDistillation}) and utilizes the incremental camera pose alignment to estimate and align the camera poses of incoming training image data within the same coordinates system as the previous camera poses (Section \ref{sec:poseSynchronization}). The overall pipeline is illustrated in Figure \ref{fig:ilNerfPipeline}.

\subsection{Replay-Based NeRF Distillation} \label{sec:knowledgeDistillation}
The problem of catastrophic forgetting occurs when a network, trained only with the currently available data at each time step, struggles to remember previously learned knowledge, resulting in low-quality image synthesis for previously seen views. To address this, we adopt a replay-based NeRF distillation strategy in the NeRF training. At each time slot $t$, we copy and freeze the parameters of the network as a teacher network before training the network on the incoming image data chunk $t$. Since the network has been trained on $t - 1$ previous image data chunks, we use $\Theta^{*}_{t-1}$ to denote the frozen parameters of the teacher network. During each training iteration for the image data chunk $t$, we use the past camera poses $\mathcal{P}^p$ to obtain the pixel colors of the past training rays from the teacher network $\Theta^{}_{t-1}$ as follows:
\begin{align}
    C^p = \mathcal{F}(\mathcal{P}^p, \Theta^{*}_{t-1})
\end{align}
By treating $C^p$ as pseudo ground truth, we optimize $\Theta_t$ by learning new knowledge from the new incoming image data and old knowledge from the teacher network, thus mitigating the forgetting problem. The new loss function is defined as follows:
\begin{align}
    & \mathcal{L} = \sum_{\textbf{r}^c}\parallel \hat{C}^c - C^c \parallel^2_2 + \sum_{\textbf{r}^p} \parallel \hat{C}^p - C^p \parallel^2_2 \\
    & \Theta_t^* = \arg \min_{\Theta_t} \mathcal{L}(\hat{C}^c, \hat{C}^p \mid C^c, \mathcal{P}^c, \Theta_{t-1}^*, \mathcal{P}^p)
\end{align}
where $\textbf{r}^c$ and $\textbf{r}^p$ are the training rays obtained from current camera poses $\mathcal{P}^c$ and past camera poses $\mathcal{P}^p$. $\hat{C}^c$ and $\hat{C}^p$ are the estimated colors by the network that is being trained, given the concatenated camera poses $[\mathcal{P}^c, \mathcal{P}^p]$:
\begin{align}
    [\hat{C}^c, \hat{C}^p] = \mathcal{F}([\mathcal{P}^c, \mathcal{P}^p], \Theta_{t})
\end{align}

The existing works \cite{chung2022meil,guo2022incremental,po2023instant} assume that the camera poses are provided with each incoming image data and not saved on the device when the next round of training image data arrives. 
This requires additional effort to train a new auxiliary neural network to remember or filter the previously trained rays. 
\ul{However, this paper considers a different scenario. Specifically, the incoming image data chunk only includes the training images and not the camera poses, and hence the camera poses need to be obtained using camera calibration techniques and saved on the device.} It's worth noting that only the previous camera poses are stored, not the previous training images. Furthermore, the camera pose requires only a small storage space. In the following section, we will discuss how to estimate and align the previous and new camera poses into the same coordinate system.

\subsection{Incremental Camera Pose Alignment} \label{sec:poseSynchronization}
% Different than the existing works, we consider a scenario in which each arrival of the image data has no camera poses. Therefore we need to get the camera poses of the arriving image data first. It cannot be simply assumed that the problem can be solved by estimating the camera poses of the arriving image data individually, because the results obtained by estimating the camera poses in isolation are not in the same coordinate system as the previous camera poses, thereby creating the coordinate shifting problem. 
Let $\mathcal{P}^p$ represent the previously aligned camera poses.
which includes intrinsic camera matrix $K$ and extrinsic camera matrices $\{\pi^p, \psi^p\}$, where $\pi^p$ is the set of rotation matrices and $\psi^p$ is the set of translations. 
When it comes to estimating camera poses from the incoming image data, it is not sufficient to assume that the problem can be resolved by independently estimating the camera poses of the incoming image data. This is because the outcomes derived from the isolated estimation of camera poses do not align with the coordinate system of the previous camera poses, leading to a significant issue of coordinate misalignment. To this end, we introduce an incremental camera pose alignment method leveraging data from a trained NeRF. We start by choosing $D$ camera poses from prior instances based on low training losses and comprehensive coverage. These are then combined with incoming images to enhance camera estimation.

% Addressing the issue of coordinate misalignment in camera pose estimation from incoming image data requires more than treating each instance independently, since isolated estimations result in discrepancies with previous camera pose coordinates. To this end, we introduce an incremental alignment method leveraging data from a trained NeRF. We start by choosing $D$ camera poses from prior instances based on low training losses and comprehensive coverage. These are then combined with incoming images to enhance camera estimation.
% When it comes to estimating camera poses from the incoming image data, it is not sufficient to assume that the problem can be resolved by independently estimating the camera poses of the incoming image data. This is because the outcomes derived from the isolated estimation of camera poses do not align with the coordinate system of the previous camera poses, leading to a significant issue of coordinate misalignment.
% To address this challenge, we propose an incremental camera pose alignment method built upon the previously generated sets of image data rendered from the trained NeRF. Initially, we select $D$ camera poses from the pool of previous camera poses based on their minimal training losses and maximal coverage of views. Then, these $D$ camera-rendered images from NeRF are integrated with the incoming image data for camera parameter estimation. 

\textbf{Finding Previous Optimal Camera Poses.}
Our approach is to formulate a reward-collection optimization problem on a graph. In this graph, the nodes represent camera positions (i.e., the translations in the camera poses) with each node assigned a reward corresponding to the negative value of the preceding training loss. The edges represent Euclidean distances between each camera pair's positions. The goal is to find a path that collects as much reward as possible, subject to constraints on the total number of visited nodes and camera view coverage. Concretely, the objective optimization problem can be formulated as
\begin{align}
& \max \sum_{k = 1}^{|\mathcal{P}^p|} x_k R_k; ~~~~s.t. ~ \sum_{k = 1}^{|\mathcal{P}^p|} x_k = D; \label{eq:numberConstraint}\\
&~~~~ S(K) \ge S_{th}, K = \{k | x_k = 1, k = 1, \dots, |\mathcal{P}^p|\} ;\label{eq:coverageConstraint}\\
&~~~~ E(x_k) \le 1, \forall k \in \{1, \dots, |\mathcal{P}^p|\} ;\label{eq:onceConstraint}
\end{align}
where $x_k$ is the binary decision variable: $x_k = 1$ if node $k$ is visited otherwise $x_k = 0$. $S(K)$ is the shortest path that connects all the selected nodes. $E(x_k)$ is the number of incoming edges of each selected node. The first constraint \eqref{eq:numberConstraint} makes sure that only $D$ previous camera poses are selected. The second constraint \eqref{eq:coverageConstraint} means that the view coverage of the selected cameras is larger than a threshold, because a large field of view coverage of the selected cameras improves the accuracy of the camera pose estimation. The third constraint \eqref{eq:onceConstraint} guarantees that every node only has one incoming edge. In other words, every node is visited at most once. 
% Note that the starting point has an incoming edge by default and the length of this edge is $0$.
Consequently, this reward-collection optimization problem can be viewed as a hybrid of the Knapsack Problem and the Travelling Salesperson Problem, which is an NP-hard problem. To solve this problem, we propose a greedy algorithm that can reduce computation time by several orders of magnitude compared with the Brute-Force method.
Let $e_{i,j} = e_{j,i}$ denote the edge between camera $i$ and $j$. At camera node $i$, we define an approximation edge length of all the nodes connecting to the node $i$ as $\hat{e}_{i, j} = R_j + \lambda(\frac{S_{th}}{D} - {e}_{i, j})$, where $\lambda$ is a parameter for adjusting the units of $R_i$ and $\frac{S{th}}{D} - {e}_{i, j}$. This approximation edge is similar to the Lagrange multiplier \cite{bertsekas2014constrained} for handling the constraint \eqref{eq:coverageConstraint}. We introduce an auxiliary starting
node into the graph, which connects to all nodes with the same edge length. The greedy algorithm begins at this auxiliary starting
node and selects the unvisited node with the maximum approximation edge as the next starting node. This process is repeated until a total of $D$ nodes have been selected.
\ul{For a comprehensive understanding of the process, we offer a detailed description of the greedy algorithm, along with pseudocode, in the} \underline{\textcolor{blue}{supplementary material}.}

% Roughly speaking, the greedy algorithm relaxes the view coverage constraint \eqref{eq:coverageConstraint} into an approximation edge, and traverses all the nodes with maximum of this edge until selecting a total $D$ camera poses.

% Roughly speaking, the greedy algorithm
% selects camera poses from the previously aligned poses in order of marginal reward, ensuring compliance with the mutual exclusion constraint, until no more elements can be selected. For a comprehensive understanding of the process, we offer an in-depth analysis of the greedy algorithm, along with pseudocode, in the supplementary material.

\textbf{Camera Pose Alignment.} After solving the above reward-collection optimization problem, we select $D$ camera poses with large view coverage from the previously aligned camera poses. 
The reason for selecting $D$ camera poses instead of using all the camera poses is that camera poses with poorly rendered images will provide inaccurate features leading to large camera estimation errors.

% choosing rendered images with low training loss provides more features to be used for feature matching, while a larger field of view allows for better estimation of the incoming image data into the original camera coordinate system. Camera poses with poorly rendered images cannot be considered because they will provide inaccurate features leading to large camera estimation errors. 

These $D$ camera poses are utilized to render the images from the NeRF model, which are integrated with the incoming image data as a group. The camera poses of this group can be estimated using camera calibration methods, such as traditional SfM or SLAM techniques. 
% Here, we use the COLMAP \cite{schonberger2016structure} to acquire the camera poses.
Let $\pi_d$ be the rotation matrix of the selected camera in time slot $t-1$ and $\tilde{\pi}_d$ be the corresponding rotation matrix of the selected camera in time slot $t$.  Similarly, let $\psi_d$ be the  translation of selected camera in time slot $t-1$ and $\tilde{\psi}_d$ be the corresponding
translation of selected camera in time slot $t$.
We can then get the transfer matrices of the rotation matrix and translation from the coordinate system in time slot $t$ to the coordinate system in time slot $t-1$ by:
\begin{align}
\resizebox{.88\hsize}{!}{$
    \triangle \pi = \frac{1}{D}\sum_{d=1}^D \pi_d (\tilde{\pi}_d)^{-1},
    \triangle \psi = \frac{1}{D}\sum_{d=1}^D (\psi_d - \triangle \pi \tilde{\psi}_d)
$}
\end{align}
% \begin{align}\label{eq:transferMatrix}
%     & \triangle \pi = \frac{1}{D}\sum_{d=1}^D \pi_d (\tilde{\pi}_d)^{-1}  \\
%     & \triangle \psi = \frac{1}{D}\sum_{d=1}^D (\psi_d - \triangle \pi \tilde{\psi}_d)
% \end{align}
We can use transfer matrices to align the camera poses $\{\pi, \psi\}$ of the new images to the original camera poses by:
\begin{align} \label{eq:synchronizePose}
    \tilde{\pi} = \triangle \pi \pi, ~~~~ \tilde{\psi} = \triangle \pi \psi + \triangle \psi
\end{align}
% \begin{align}\label{eq:synchronizePose}
%     & \tilde{\pi} = \triangle \pi \pi\\
%     & \tilde{\psi} = \triangle \pi \psi + \triangle \psi
% \end{align}

\textbf{Joint Optimization of Poses and NeRF.}\label{sec:poseRefinement}
However, we have observed that the camera pose alignment still produces errors in the camera poses. To address this, we draw inspiration from previous works \cite{wang2021nerf,jeong2021self} and introduce a joint optimization of poses and NeRF method during training of our proposed IL-NeRF. 
Rather than directly optimizing the initial camera pose $\tilde{\pi}$ and $\tilde{\psi}$, we employ a 6-dimensional vector $\Phi = [a, b]$ to define the trainable parameters of each camera pose, where $a \in \mathbb{R}^3$ represents the 3D rotation angles and $b \in \mathbb{R}^3$ denotes the increment of the translation.
To ensure the orthogonality of the rotation matrix, Rodrigues' formula $\Omega(a)$ is used to generate the 3D rotation matrix. Final rotation and translation are:
\begin{align} \label{eq:poseRefinement}
    \tilde{\pi} = \Omega(a) \tilde{\pi}, ~~~~ \tilde{\psi} = \tilde{\psi} + b
\end{align}
% \begin{align}\label{eq:poseRefinement}
%     & \tilde{\pi} = \Omega(a) \tilde{\pi}\\
%     & \tilde{\psi} = \tilde{\psi} + b
% \end{align}
By integrating the 6-dimensional vector $\Phi$ into the NeRF training pipeline, the camera parameters and scene representation can be jointly optimized during training. Here, we slightly abuse the notation to use $\Phi$ to represent the group of all cameras' trainable parameters. Mathematically, the pose refinement can be written as:
\begin{align}
    \Theta_t^*, \Phi_t^*  = \arg \min_{\Theta_t, \Phi_t} \mathcal{L}(\hat{C}^c, \hat{C}^p \mid C^c, \mathcal{P}^c, \Theta_{t-1}^*, \mathcal{P}^p)
\end{align}

\vspace{-0.3cm}
% The pipeline of IL-NeRF is summarized in Algorithm \ref{alg:alorithm1}.

\begin{algorithm}
	\caption{IL-NeRF Pseudo Code}
	\begin{algorithmic}[1]
		\State Initialize $\mathcal{P}^p =  \emptyset$.
		\For {$t=0$}
        \State Estimate camera poses $\mathcal{P}^c_0$ from $G^0$
		\State Jointly train NeRF network $\Theta_0$ with camera poses
        \State $\mathcal{P}^p =  \mathcal{P}^p \bigcup \mathcal{P}^c_0$ 
		\EndFor
		\For {$t=1$ to $t = T$}
		\State Copy and freeze as $\Theta_{t-1}^*$
        \State Obtain past training data $C^p$ by $\mathcal{F}(\mathcal{P}^p, \Theta^{*}_{t-1})$
        \State Align the camera pose  $\mathcal{P}^c_t$ based on $\mathcal{P}^p$
		\State Jointly train NeRF network $\Theta_t$ with camera poses
        \State $\mathcal{P}^p =  \mathcal{P}^p \bigcup \mathcal{P}^c_t$ 
		\EndFor
	\end{algorithmic}
	\label{alg:alorithm1}
	% \vspace{-10 pt}
\end{algorithm}
%------------------------------------------------------------------------
\vspace{-10 pt}
\section{Experiment} \label{sec:experiment}
\textbf{Dataset.}
We use three different datasets to evaluate different aspects of our model, namely Mip-NeRF360 \cite{barron2022mip}, LLFF \cite{mildenhall2019local}, and NeRF-real360 \cite{mildenhall2021nerf}. 
To simulate the incremental scenarios, we reorganize the camera order so that it moves sequentially and we select a portion of the dataset where the previous images are not revisited. 
All the training images of each scene are divided into four training chunks denoted as $\mathcal{G} = \{G^0, G^1, G^2, G^3 \}$.
We acquire the camera pose parameters using COLMAP \cite{schonberger2016structure}. 

\begin{table*}[!htbp]
\centering
	\caption{Performance comparison with the baselines on PSNR, SSIM, and LPIPS.  IL-NeRF outperforms the original NeRF, EWC, NeRF-SLAM and
achieves comparable results with NeRF$^*$. Note that NeRF$^*$, NeRF, and EWC require the ground truth pre-estimated camera poses from entire image data, but IL-NeRF estimates and aligns camera poses by the proposed incremental camera pose alignment module. \textbf{NeRF$^*$ can be treated as the representation of the existing incremental learning works with accurate camera poses~\cite{chung2022meil,guo2022incremental,po2023instant,cai2023clnerf}.}}
\vspace{-0.3cm}
    \begin{tabular}{p{0.2cm}p{0.2cm} p{2.3cm}|c|cccc}
    \toprule[1.5pt]
        \multirow{2}{*}{Data} & \multirow{2}{*}{} &  \multirow{2}{*}{Method} & Need & \multicolumn{4}{c}{\textbf{PSNR $\Uparrow$ / SSIM $\Uparrow$ / LPIPS $\Downarrow$}}\\
        \cline{5-8}
        & & & Pose & $G^0$ & $G^1$ & $G^2$ & $G^3$\\
        \hline
        \multirow{8}{*}{\begin{turn}{-90}Mip-NeRF360\end{turn}} & \multirow{4}{*}{\begin{turn}{-90}Counter\end{turn}} & NeRF$^*$ & Yes & \cellcolor{gray!25}32.17 / 0.92 / 0.07 & \cellcolor{gray!25}29.58 / 0.86 / 0.14 & \cellcolor{gray!25}28.03 / 0.82 / 0.18 & \cellcolor{gray!25}28.28 / 0.85 / 0.18 \\
        & & NeRF & Yes & 32.12 / 0.91 / 0.07 & 24.62 / 0.72 / 0.25 & 21.94 / 0.65 / 0.34 & 20.30 / 0.62 / 0.37 \\
        & & EWC & Yes & 32.12 / 0.91 / 0.07 & 23.83 / 0.72 / 0.25 & 22.56 / 0.65 / 0.33 & 21.11 / 0.61 / 0.36 \\
        % \cline{2-7}
        & & NeRF-SLAM & No & 31.75 / 0.91 / 0.07 & 28.30 / 0.83 / 0.21 & 26.84 / 0.79 / 0.28 & 25.30 / 0.77 / 0.31 \\  
        & & IL-NeRF (Ours) & No & 32.13 / 0.91 / 0.07
        & \textbf{29.63} / \textbf{0.87} / \textbf{0.12} & \textbf{28.56} / \textbf{0.85} / \textbf{0.15} & \textbf{27.82} / \textbf{0.83} / \textbf{0.17} \\ 
        \cline{3-8}
        & \multirow{4}{*}{\begin{turn}{-90}Kitchen\end{turn}}& NeRF$^*$ & Yes & \cellcolor{gray!25}31.05 / 0.91 / 0.07 & \cellcolor{gray!25}29.72 / 0.88 / 0.13 & \cellcolor{gray!25}29.33 / 0.85 / 0.15 & \cellcolor{gray!25}29.18 / 0.84 / 0.14 \\
        & & NeRF & Yes & 31.17 / 0.91 / 0.08 & 27.01 / 0.75 / 0.25  & 21.42 / 0.70 / 0.31 & 23.69 / 0.75 / 0.24\\
        & & EWC & Yes & 31.17 / 0.91 / 0.08 & 26.76 / 0.74 / 0.25 & 22.09 / 0.70 / 0.31 & 23.39 / 0.74 / 0.23 \\
        & & NeRF-SLAM & No & 30.87 / 0.90 / 0.09 & 29.63 / 0.85 / 0.20 & 27.65 / 0.81 / 0.24 & 27.71 / 0.82 / 0.20 \\
        & & IL-NeRF (Ours) & No & 31.27 / 0.92 / 0.07
        & \textbf{30.66} / \textbf{0.89} / \textbf{0.10} & \textbf{29.84} / \textbf{0.87} / \textbf{0.12} & \textbf{29.34} / \textbf{0.86} / \textbf{0.13} \\
        \hline
        \multirow{8}{*}{\begin{turn}{-90} LLFF \end{turn}} & \multirow{4}{*}{\begin{turn}{-90}Fortress\end{turn}}& NeRF$^*$ & Yes & \cellcolor{gray!25}31.75 / 0.86 / 0.11 & \cellcolor{gray!25}31.83 / 0.84 / 0.09 & \cellcolor{gray!25}30.90 / 0.86 / 0.14 & \cellcolor{gray!25}29.81 / 0.85 / 0.11 \\
        & & NeRF & Yes & 31.56 / 0.85 / 0.11 & 29.38 / 0.80 / 0.15 & 27.05 / 0.78 / 0.17 & 25.39 / 0.78 / 0.16\\
        & & EWC & Yes & 31.56 / 0.85 / 0.11 & 29.41 / 0.79 / 0.15 & 25.83 / 0.77 / 0.16 & 24.40 / 0.78 / 0.15\\
        & & NeRf-SLAM & No & 31.08 / 0.82 / 0.12 & 31.09 / 0.82 / 0.12 & 29.77 / 0.83 / 0.15 & 28.53 / 0.82 / 0.14\\
        & & IL-NeRF & No & 31.69 / 0.85 / 0.11 & \textbf{31.02} / \textbf{0.84} / \textbf{0.10} & \textbf{30.33} / \textbf{0.84} / \textbf{0.11} & \textbf{29.45} / \textbf{0.83} / \textbf{0.12}\\ 
         \cline{3-8}
        & \multirow{4}{*}{\begin{turn}{-90}Horns\end{turn}}& NeRF$^*$ & Yes & \cellcolor{gray!25}29.86 / 0.89 / 0.07 & \cellcolor{gray!25}29.67 / 0.89 / 0.06 & \cellcolor{gray!25}29.24 / 0.89 / 0.07 & \cellcolor{gray!25}28.87 / 0.87 / 0.08\\
        & & NeRF & Yes & 29.78 / 0.86 / 0.09 & 27.04 / 0.75 / 0.09 & 26.04 / 0.74 / 0.12 & 24.01 / 0.70 / 0.14\\
        & & EWC & Yes & 29.78 / 0.86 / 0.09 & 26.77 / 0.75 / 0.09 & 27.08 / 0.74 / 0.11 &  24.68 / 0.69 / 0.14 \\
        & & NeRF-SLAM & No & 28.86 / 0.83 / 0.10 & 28.77 / 0.84 / 0.08 & 28.19 / 0.83 / 0.10 &  27.56 / 0.82 / 0.12 \\
        & & IL-NeRF (Ours)& No & 29.92 / 0.89 / 0.07 & \textbf{29.50} / \textbf{0.89} / \textbf{0.07} & \textbf{29.01} / \textbf{0.89} / \textbf{0.07} & \textbf{28.96} / \textbf{0.87} / \textbf{0.09}\\
        \hline
        \multirow{8}{*}{\begin{turn}{-90} NeRF-real360 \end{turn}} & \multirow{4}{*}{\begin{turn}{-90}Pinecone\end{turn}}&  NeRF$^*$ & Yes & \cellcolor{gray!25}26.88 / 0.89 / 0.12 & \cellcolor{gray!25}24.23 / 0.79 / 0.16 & \cellcolor{gray!25}24.03 / 0.73 / 0.19 & \cellcolor{gray!25}23.18 / 0.74 / 0.21 \\
        & & NeRF & Yes & 26.22 / 0.84 / 0.16 & 22.90 / 0.64 / 0.24 & 21.15 / 0.58 / 0.33 & 18.94 / 0.49 / 0.41 \\
        & & EWC & Yes & 26.22 / 0.84 / 0.16 & 22.70 / 0.63 / 0.24 & 21.42 / 0.57 / 0.32 & 18.81 / 0.48 / 0.41 \\
        & & NeRF-SLAM & No & 25.63 / 0.81 / 0.18 & 24.09 / 0.73 / 0.22 & 23.01 / 0.68 / 0.29 & 21.79 / 0.65 / 0.34 \\
        & & IL-NeRF (Ours) & No & 26.3 / 0.87 / 0.10
        & \textbf{24.56} / \textbf{0.78} / \textbf{0.17} & \textbf{23.78} / \textbf{0.74} / \textbf{0.20} & \textbf{22.93} / \textbf{0.72} / \textbf{0.23} \\
         \cline{3-8}
        & \multirow{4}{*}{\begin{turn}{-90}Vasedeck\end{turn}}& NeRF$^*$ & Yes & \cellcolor{gray!25}29.27 / 0.86 / 0.07 & \cellcolor{gray!25}27.93 / 0.85 / 0.12 & \cellcolor{gray!25}26.03 / 0.74 / 0.16 & \cellcolor{gray!25}26.18 / 0.74 / 0.18 \\
        & & NeRF & Yes & 29.03 / 0.85 / 0.07 & 23.99 / 0.70 / 0.26 & 22.73 / 0.69 / 0.24 & 21.57 / 0.64 / 0.31 \\
        & & EWC & Yes & 29.03 / 0.85 / 0.07 & 24.36 / 0.69 / 0.25 & 22.25 / 0.68 / 0.24 & 20.52 / 0.64 / 0.30 \\
        & & NeRF-SLAM & No & 27.98 / 0.79 / 0.11 & 26.41 / 0.77 / 0.21 & 25.10 / 0.72 / 0.21 & 24.62 / 0.71 / 0.26 \\
        & & IL-NeRF (Ours) & No & 29.48 / 0.86 / 0.07
        & \textbf{27.38} / \textbf{0.82} / \textbf{0.10} & \textbf{26.11} / \textbf{0.76} / \textbf{0.14} & \textbf{26.15} / \textbf{0.75} / \textbf{0.17} \\
        \toprule[1.5pt]
    \end{tabular}
\label{table:accuracyPerformanceComparison}
\vspace{-10 pt}
\end{table*}

\textbf{Baseline and Metrics.}
IL-NeRF is compared with the following baselines: \textbf{NeRF}: Original NeRF is incrementally trained with only current image data chunk with ground truth camera poses but without NeRF distillation, making it susceptible to catastrophic forgetting. \textbf{EWC} \cite{kirkpatrick2017overcoming}: Similar to the NeRF, EWC incrementally trains the model with only current image data chunk with ground truth camera poses however it utilizes a widely-used regularization-based method, which penalizes the changes in parameters that are important for past
training sets. \textbf{NeRF$^*$}: NeRF is incrementally trained with ground truth camera poses under replay-based NeRF distillation. Note that the ground truth camera poses are estimated from all the training images. \textit{NeRF$^*$ can be treated as the representation of the existing works \cite{chung2022meil,guo2022incremental,po2023instant,cai2023clnerf}, which require the ground truth camera poses in each incoming image data chunk.} \textbf{NeRF-SLAM}: we follow the general implementation of NeRF-SLAM \cite{rosinol2022nerf}, which use the SLAM to align camera poses of the coming image data. We replace Droid-SLAM \cite{teed2021droid} in the NeRF-SLAM to ORB-SLAM2 \cite{mur2017orb} because Droid-SLAM utilizes a complex deep learning model for camera pose estimation, which needs training on image data before NeRF training.

% however, it requires memory scalability for storing the point cloud and key frames for PnP algorithm.

We evaluate IL-NeRF and baselines in three aspects, including
Peak Signal-to-Noise Ratio (PSNR), Structural Similarity
Index Measure (SSIM) \cite{wang2004image} and Learned Perceptual Image
Patch Similarity (LPIPS) \cite{zhang2018unreasonable}. We use AlexNet \cite{krizhevsky2017imagenet} as the backbone of LPIPS. It should be noted that as joint optimization of poses and NeRF is performed for all camera poses, including the previous and current poses, at each time step, the camera poses are continuously changing. This means it is not possible to fix the camera poses for test data, and all evaluation metrics compare the rendered images from the training camera poses with the ground truth.

% \subsection{Implementation Details}
% We implement our framework following the architecture of Instant-NeRF \cite{muller2022instant,ngp-pl}. We use two separate Adam optimizers for NeRF and camera poses refinement respectively, with an initial learning rate of 0.01 for NeRF and an initial learning rate of 0.005 for pose refinement. The learning rate of the NeRF model decays every iteration by multiplying with 0.9954 (exponential decay), and the learning rate of the pose refinement decays every 100 iterations with a multiplier of 0.9. We train the network in each incremental step for Mip-NeRF360 with 30k iterations and $D=10$, LLFF with 5k iterations and $D=8$, and NeRF-real360 with 20k iterations and $D=10$. 

\begin{figure*}[!htbp]
	\centering
	\includegraphics[width=\linewidth]{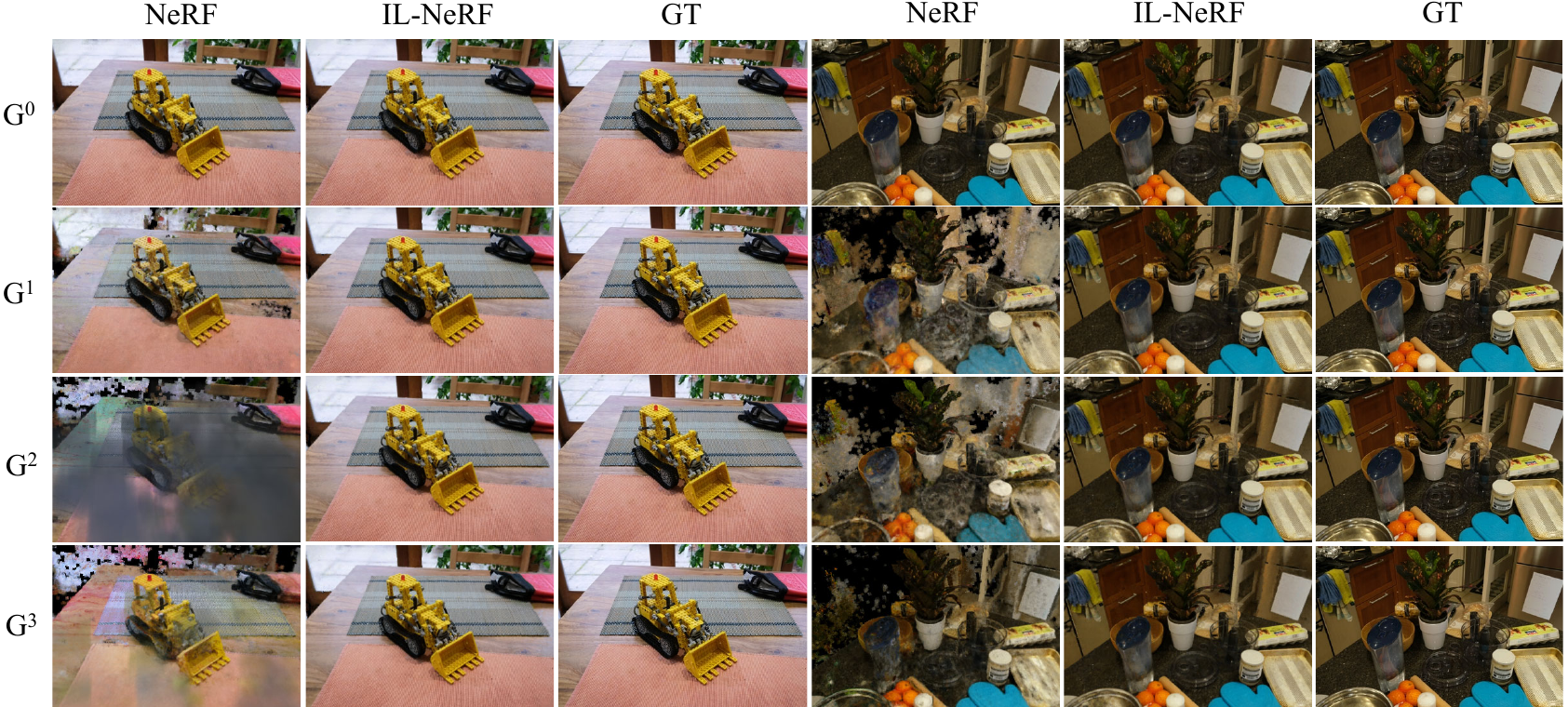}
    \vspace{-20 pt}
	\caption{Qualitative comparison of the original NeRF and IL-NeRF on the rendering images in the first image data after each incremental training. GT means the ground truth of the training image. The original NeRF demonstrates severe catastrophic forgetting, leading to the loss of early-task scene information. In contrast, IL-NeRF is able to preserve the scene of interest throughout the training process.}
	\label{fig:visualization}
	\vspace{-15 pt}
\end{figure*}

\subsection{Results}
% Table \ref{table:accuracyPerformanceComparison} shows the results obtained by IL-NeRF and baseline methods on the three datasets, e.g., `Counter' and `Kitchen' in Mip-NeRF360 dataset, `Fortress' and `Horns' in LLFF dataset, `Pinecone' and `Vasedeck' in NeRF-real360 dataset. Due to page limitations, more comparisons are shown in the supplementary material.
% From the results, we can see that IL-NeRF outperforms the original NeRF and achieves comparable results with NeRF$^*$. We explain the results in more detail next.

Table \ref{table:accuracyPerformanceComparison} shows the partial results obtained by IL-NeRF and baseline methods on the three datasets. Due to page limitations, \textit{more comparisons are shown in the supplementary material}.
From the results, we can see that IL-NeRF outperforms the original NeRF, EWC, NeRF-SLAM and achieves comparable results with NeRF$^*$. We explain the results in more detail next.

\textbf{NeRF:} In comparison to the original NeRF, IL-NeRF demonstrates substantial enhancements of between $10.87\%$ to $36.55\%$ in terms of PSNR, $15.18\%$ to $46.90\%$ in terms of SSIM, and $25.00\%$ to $54.05\%$ in terms of LPIPS across three datasets. In the initial image data chunk $G^0$, IL-NeRF outperforms the original NeRF slightly, as a result of joint optimization assistance provided by IL-NeRF. However, the performance of the original NeRF rapidly declines thereafter due to catastrophic forgetting.

\textbf{EWC:} EWC fails to reduce the adverse effects of catastrophic forgetting, and in some cases can be even worse than traditional NeRF. This is due to the lack of previous training images, the failure of EWC to recover previous scenes, and the introduction of a penalty mechanism that creates a disincentive to learn new images that have not been scanned.

\textbf{NeRF$^*$:} Comparing NeRF$^*$ with IL-NeRF may not be entirely fair, given that NeRF$^*$ benefits from having access to all training images to estimate camera poses for the image data chunks, while in our scenario, the camera poses are not provided in the task and must be derived through incremental camera pose alignment. Nonetheless, IL-NeRF performs comparably to NeRF$^*$, largely due to its incremental camera pose alignment module utilized during training.

\textbf{NeRF-SLAM:} While NeRF-SLAM uses SLAM to align camera poses in incoming image data to a common coordinate system, it lags behind our IL-NeRF in terms of performance. This difference stems from NeRF-SLAM's exclusive reliance on selected keyframes for replay-based training, resulting in overfitting to specific rays and compromising multiview consistency. Furthermore, NeRF-SLAM demands more memory storage space. For instance, on the `Garden' image data in Mip-NeRF360, NeRF-SLAM necessitates an additional 251.3 MB of memory for storing keyframes and point clouds. In contrast, IL-NeRF only requires an additional 37.7 KB of memory for storing previous camera poses.

% \begin{table}[!ht]
% \small
% \centering
% 	\caption{Addition Memory Usage of IL-NeRF and baselines.}
%     \begin{tabular}{l|c|c|c}
%     \toprule[1.5pt]
%     Scene & NeRF/EWC & NeRF$^*$ / IL-NeRF  & NeRF-SLAM  \\
%     \cline{1-4}
%     Kitchen & 0 & 37.7 KB & 251.3 MB  \\
%     Vasedeck & 0 & 15.8 KB & 202.6 MB  \\
%     \toprule[1.5pt]
%     \end{tabular}
% \label{table:additionMemory}
% \end{table}

% Table \ref{table:additionMemory} shows the additional memory storage for IL-NeRF and baselines.

Figure \ref{fig:visualization} provides additional insights by presenting a qualitative comparison of the performance of the original NeRF and IL-NeRF on the `Kitchen' and `Counter' scenes in the Mip-NeRF360 dataset. Specifically, we demonstrate the rendering results on the first image data after each incremental training. It is evident that the original NeRF suffers from the catastrophic forgetting problem, resulting in images with significant distortions such as noise and blur, whereas IL-NeRF generates highly realistic images with quality comparable to the ground truth. This observation indicates that IL-NeRF is highly effective in mitigating the forgetting problem and estimating the camera poses. \textit{More results are shown in the supplementary material}.

\begin{figure}[tt]
	\centering
	\includegraphics[width=\linewidth]{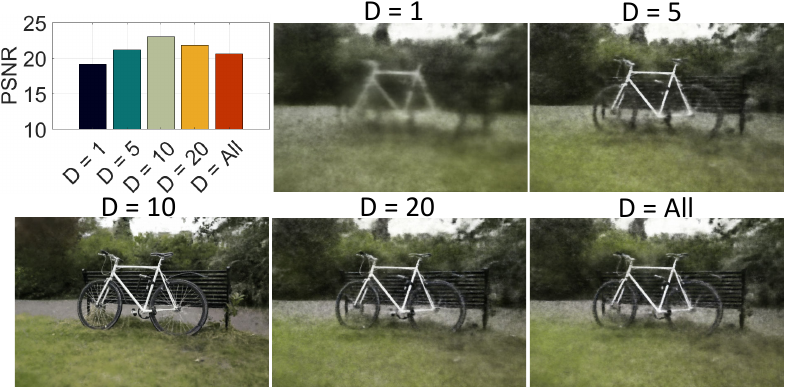}
    \vspace{-20 pt}
	\caption{Influence of optimal pose count D on IL-NeRF.}
	\label{fig:imapctD}
	\vspace{-15 pt}
\end{figure}

\subsection{Ablation Study}
\begin{figure}[tt]
	\centering
	\begin{minipage}[t]{0.45\linewidth}
		\includegraphics[width=\textwidth]{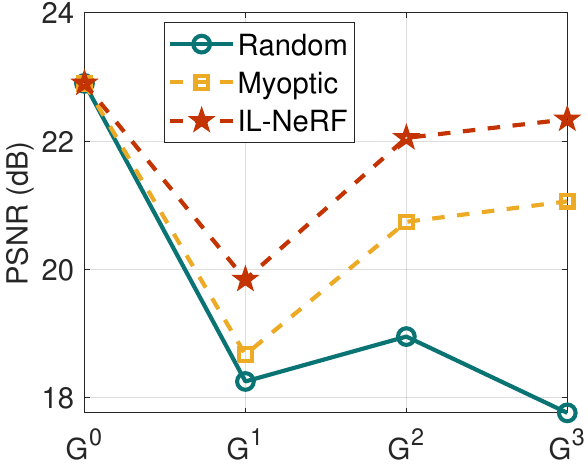}
		\vspace{-20 pt}
		\caption{Our camera pose selection method outperforms random selection and myopic selection.}
		\label{fig:compareSelectionMethods}
	\end{minipage}
	\hspace{0.03\linewidth}
	\begin{minipage}[t]{0.46\linewidth}
		\includegraphics[width=\textwidth]{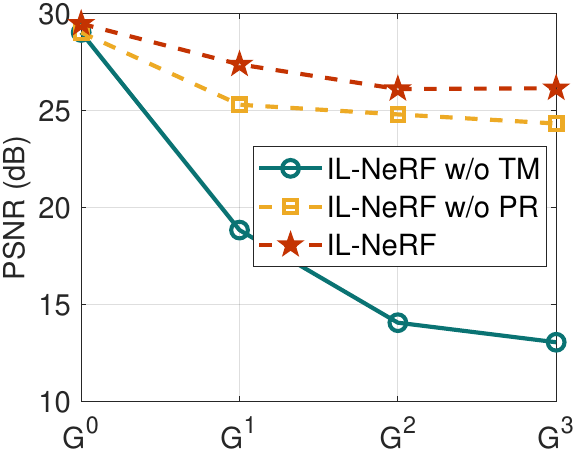}
		\vspace{-20 pt}
		\caption{Comparison of IL-NeRF w/o TM, IL-NeRF w/o PR and IL-NeRF. IL-NeRF outperforms these two cases.}
		\label{fig:impactTMandPR}
	\end{minipage}
	% \vspace{-10 pt}
\end{figure}

\begin{figure}[tt]
	\centering
	\includegraphics[width=\linewidth]{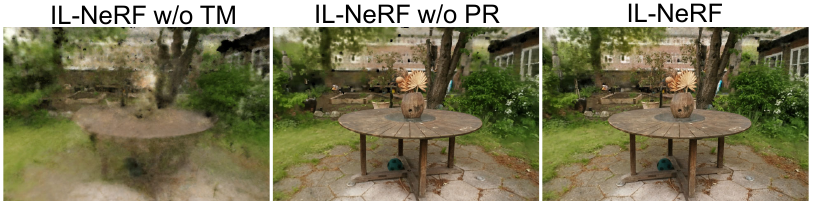}
    \vspace{-20 pt}
	\caption{Comparison of IL-NeRF w/o Transfer Matrices (TM), IL-NeRF w/o Pose Refinement (PR) and IL-NeRF.}
	\label{fig:qualitativeCompareTMandpr}
	\vspace{-10 pt}
\end{figure}

\begin{figure}
  \centering
  \begin{subfigure}{0.48\linewidth}
    % \fbox{\rule{0pt}{2in} \rule{.45\linewidth}{0pt}}
    \includegraphics[width=\linewidth]{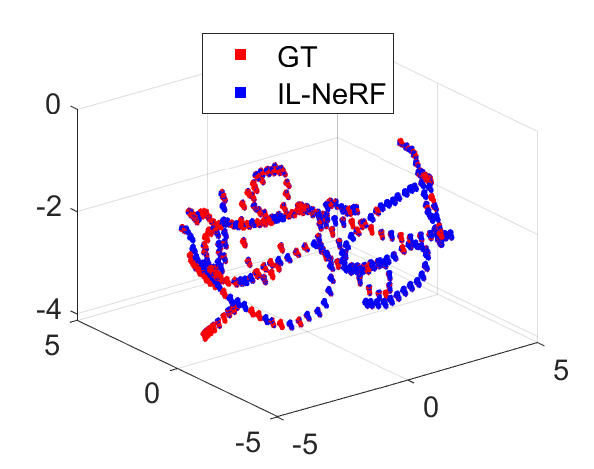}
    \caption{Kitchen}
    \label{fig:KitchenCameraPose}
  \end{subfigure}
  % \hfill
  \begin{subfigure}{0.48\linewidth}
    % \fbox{\rule{0pt}{2in} \rule{.45\linewidth}{0pt}}
    \includegraphics[width=\linewidth]{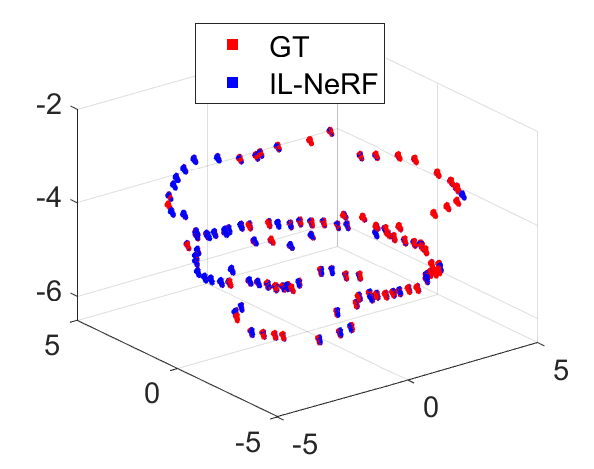}
    \caption{Vasedesk}
    \label{VasedeskCameraPose}
  \end{subfigure}
  \vspace{-10 pt}
  \caption{Camera pose estimation comparison. GT means the camera poses estimated by COLMAP from all the training images. IL-NeRF recovers accurate camera poses with the help of incremental camera pose alignment.}
  \label{fig:cameraPose}
  \vspace{-15 pt}
\end{figure}

\textbf{Effect of Pose Selection.} The first step of incremental camera pose alignment is to find previous $D$ optimal camera poses as the references for estimating and aligning the camera poses of the incoming image data. 
To illustrate the influence of $D$ on the IL-NERF performance, we set the value of $D$ to 1, 5, 10, 20, and all, respectively. Figure \ref{fig:imapctD} portrays the PSNR of IL-NeRF across varying values of $D$ on the `Bicycle' scene from the Mip-NeRF360 dataset. As depicted, when $D$ is a small value, the lack of adequate reference images results in the estimated camera poses of the incoming image data deviating from the original camera pose coordinate system, thereby leading to considerably poor rendering. As the value of $D$ increases, the estimated camera poses of the incoming image data become increasingly precise, and thus the PSNR increases. Note that an excessively large value of $D$ introduces poorly rendered cameras, subsequently leading to a decrease in PSNR. To identify the optimal $D$ camera poses, we address a reward-collection optimization problem on a graph. In order to demonstrate the effectiveness of our camera selection approach, we conduct a comparative analysis with two baselines: (1) Randomly selecting $D$ camera poses as the reference, and (2) Myopically selecting $D$ camera poses with the lowest training losses. Figure \ref{fig:compareSelectionMethods} shows the performance of IL-NeRF and two baselines on the `Bicycle' scene. As we can see, our proposed method surpasses the other two approaches. This is primarily due to our method's ability to ensure the quality of rendered images used as references while providing a broader camera view coverage, thereby facilitating the more accurate camera pose estimation of incoming image data.\\
\textbf{Effect of Transfer Matrices.} 
The transfer matrices are obtained by computing the corresponding rotation matrix and translation of the selected D camera poses in time slots $t-1$ and $t$. These matrices are then employed to align the camera pose of new images to the original camera pose coordinate system. To investigate the effectiveness of the transfer matrices, we compare IL-NeRF with IL-NeRF without considering the transfer matrices, denoted as `IL-NeRF w/o TM' in Figure \ref{fig:impactTMandPR} on the `Garden' scene from the Mip-NeRF360 dataset. The results reveal a significant decline in performance without the transfer matrices, achieving only $15.76dB$ in terms of PSNR. This decline can be attributed to separate camera pose estimation for two tasks resulting in camera poses in two independent coordinate systems, which could mislead the model during training, leading to decreased performance. \\
\textbf{Effect of Pose Refinement.} 
Despite the transfer matrices' ability to align the camera poses to the original coordinate system, they may still contain noise and inaccuracies. To mitigate this issue, we use the coordinate-aligned camera poses as initial values and jointly optimize the camera poses and scene representation during NeRF training, a process known as pose refinement. We perform an ablation study to investigate the effectiveness of pose refinement by comparing IL-NeRF with and without it. The results in Figure \ref{fig:impactTMandPR} indicate that IL-NeRF with pose refinement outperforms IL-NeRF without it (i.e., IL-NeRF w/o PR). Figure \ref{fig:qualitativeCompareTMandpr} further shows the qualitative comparison of IL-NeRF w/o TM, IL-NeRF w/o PR, and IL-NeRF. \textit{More results are shown in the supplementary material}. \\
\textbf{Camera Pose.} Our goal is to incrementally train a NeRF model given only RGB images as input, without known camera poses. In other words, we need to find out the camera poses associated with each input image while training the NeRF model.
We treat the COLMAP estimation from all training images as ground-truth (GT) camera poses and report the difference between our optimized
camera poses and theirs on the training images. Figure \ref{fig:cameraPose} shows the camera pose trajectories of GT and IL-NeRF. As we can see, IL-NeRF recovers accurate camera poses with the help of camera coordinate alignment and pose refinement. \textit{More results are shown in the supplementary material}.

% \section{Limitation}
% For large-scale scenes with limited overlap between views in the training dataset, the performance of IL-NeRF may be suboptimal because the limited overlap between views can result in significant errors or even the inability to calculate the transfer matrices during the camera coordinate alignment.

\vspace{-5 pt}
\section{Conclusion}
In this study, we introduce an incremental learning algorithm called IL-NeRF that tackles the problems of catastrophic forgetting and coordinate shifting in NeRF training in incremental learning settings. The IL-NeRF algorithm employs a replay-based NeRF distillation pipeline to retain past information and learn from new data independently. Furthermore, a method for aligning camera pose coordinates is introduced to identify camera poses associated with incoming tasks during the NeRF model training. Experimental results reveal that the IL-NeRF algorithm outperforms the original NeRF model in sequential data settings.

\clearpage
{
    \small
    \bibliographystyle{ieeenat_fullname}
    \bibliography{main}
}

\clearpage
\setcounter{page}{1}
\maketitlesupplementary

\section{Finding Previous Optimal Camera Poses}
To find previous $D$ optimal camera poses, we formulate a reward-collection optimization problem on a graph. In this graph, the nodes represent camera positions (i.e., the translations in the camera poses) with each node assigned a reward corresponding to the negative value of the preceding training loss. The edges represent Euclidean distances between each camera pair's positions. The goal is to find a path that collects as much reward as possible, subject to constraints on the total number of visited nodes and camera view coverage. Concretely, the objective optimization problem can be formulated as
\begin{align}
& \max \sum_{k = 1}^{|\mathcal{P}^p|} x_k R_k\\ 
& s.t. ~ \sum_{k = 1}^{|\mathcal{P}^p|} x_k = D \label{eq:numberConstraint_appendex}\\
&~~~~ S(K) \ge S_{th}, K = \{k | x_k = 1, k = 1, \dots, |\mathcal{P}^p|\} \label{eq:coverageConstraint_appendex}\\
&~~~~ E(x_k) \le 1, \forall k \in \{1, \dots, |\mathcal{P}^p|\} \label{eq:onceConstraint_appendex}
\end{align}
where $x_k$ is the binary decision variable: $x_k = 1$ if node $k$ is visited otherwise $x_k = 0$. $S(K)$ is the shortest path that connects all the selected nodes. $E(x_k)$ is the number of incoming edge of each selected node. The first constraint \eqref{eq:numberConstraint_appendex} makes sure that only $D$ previous camera poses are selected. The second constraint \eqref{eq:coverageConstraint_appendex} means that the view coverage of the selected cameras is larger than a threshold. This is because a large field of view coverage of the selected cameras improves the accuracy of the camera pose estimation. The third constraint \eqref{eq:onceConstraint_appendex} guarantees that every node only has one incoming edge. In other words, every node is visited at most once. 
Consequently, this reward-collection optimization problem can be viewed as a hybrid of the Knapsack Problem and the Travelling Salesperson Problem, which is an NP-hard problem. 

\textbf{Related Work for Reward-Collection Optimization Problem.} The reward collection problem, also named orienteering problem, is an optimization issue that aims to determine the most efficient route for visiting
multiple locations while maximizing the value or score of each
place seen, all within a specified time frame and beginning and
ending at a particular point \cite{golden1987orienteering}.  This problem is widely utilized in the tourism sector \cite{vansteenwegen2011orienteering}, robot routing \cite{pvenivcka2017dubins}, food delivery \cite{vansteenwegen2019orienteering} and transportation control \cite{martins2021electric}. As the
orienteering problem belongs to the NP-hard class of
problems, no algorithm can solve it optimally within a
reasonable amount of time \cite{jandaghi2021categorized}. Different from the traditional orienteering problem, the optimization problem in this paper is more complex. Firstly, we do not limit the start and end points, and at the same time, we have a limitation on the number of accessible points, which makes it impossible to apply the existing proposed approaches to our method.

\begin{algorithm}
	\caption{Brute-Force for Selecting Cameras}
	\begin{algorithmic}[1]
		\State Generate all possible $D$ camera combinations $\mathcal{K}$.
        \State Initialize $\mathcal{B} = \emptyset$ and $b = -\infty$.
		\For {$K \in \mathcal{K}$}
        \State Use Breath-First-Search to find the shortest path $S(K)$ that visits all the nodes in $K$.
        \If {$S(K) \ge S_{th}$}
            \State Sum the total reward $\mathcal{R}$ of $K$ nodes.
            \If {$\mathcal{R} \ge b$}
                \State $\mathcal{B} = K$
                \State $b = \mathcal{R}$
            \EndIf
        \EndIf
        \EndFor
		\State Return $\mathcal{B}$
	\end{algorithmic}
	\label{alg:alorithm_bruthForce}
	% \vspace{-20 pt}
\end{algorithm}

\textbf{Brute-Force Method.} The straightforward approach to address this problem is the Brute-Force method, demonstrated in Algorithm \ref{alg:alorithm_bruthForce}. This method involves: (1) Determining the shortest path for visiting all nodes for each $D$ camera combination. (2) Selecting the camera combination with the highest total rewards, while ensuring compliance with all constraints. However, the time complexity of this approach is $O((2^D \times D) \times \binom{N}{D})$, where $\binom{N}{D}$ represents the number of $D$-combinations derived from a given set of $N$ previous camera poses. $2^D \times D$ is the time complexity of finding the shortest path for each D camera combination. While this method guarantees an optimal solution, it becomes impractical for large numbers of nodes due to its time complexity.

\textbf{Proposed Greedy Algorithm.} Let $\mathbb{G}(V, E)$ be the graph of the previous camera poses, where $V$ denotes the set of cameras (nodes) and $E$ denotes the set of edges connecting each pair of two cameras. Let $e_{i,j} = e_{j,i}, e_{i,j} \in E, e_{j,i} \in E$ denote the edge between camera $i$ and $j$.
Note that $\mathbb{G}(V, E)$ is an undirected weighted complete graph. The core concept of our greedy algorithm is to traverse all unvisited nodes starting from a specific node. During this process, the algorithm calculates the approximate edges between each pair of the current node and its connected unvisited node, subsequently selecting the node with the maximum approximation edge as the next starting node. This process is repeated until a total of $D$ nodes have been selected. The complete description of the greedy algorithm is outlined in Algorithm \ref{alg:greedy}.

\textbf{Step 1 (line 1 to line 2):} 
Introducing an auxiliary starting node $V_0$ into the graph, which establishes connections to all nodes with an edge length of 0. We define a set $\mathcal{B}$ to keep track of the visited nodes during traversal and initialize it as $\mathcal{B} = {V_0}$. Additionally, the current node index is initialized as $k = 0$.

\textbf{Step 2 (line 4 to line 10):} 
Identifying all the connected nodes of the current node $V_k$ that have not been visited yet (i.e., $V_i \notin \mathcal{B}$). Based on the reward $R_i$ and the edge length $e_{k, i}$ for each unvisited node, we compute the approximation edge length $\hat{e}_{k, i}$ and insert it into a temporary set $\hat{E}$. Specifically, $\hat{e}_{k, i} = R_i + \lambda(\frac{S_{th}}{D} - {e}_{k, i})$, where $\lambda$ is a parameter for adjusting the units of $R_i$ and $\frac{S{th}}{D} - {e}_{k, i}$. This approximation edge is similar to the Lagrange multiplier \cite{bertsekas2014constrained} for handling the constraint \eqref{eq:coverageConstraint_appendex}.

\textbf{Step 3 (line 11 to line 12):}
Selecting the node with the maximum $\hat{e}_{k, i}$ as the next visited node, updating the current index as $k = \arg \max \hat{E}$, and inserting the next visited node into the set of visited nodes $\mathcal{B}$.

\textbf{Step 4:} Repeating Step 2 and Step 3 until the greedy algorithm has visited a total of $D$ nodes.

\begin{algorithm}
	\caption{Proposed Greedy Algorithm}
	\begin{algorithmic}[1]
		\State Add an auxiliary starting node $V_0$ linking all the nodes.
        \State Initialize the visited set $\mathcal{B} = \{V_0\}$ and $k = 0$.
        \While {$|\mathcal{B}| < D + 1$}
        \State $\hat{E} = \{\}$
            \For {$V_i \in V$}
                \If {$V_i \notin \mathcal{B}$}
                    \State $\hat{e}_{k, i} = R_i + \lambda(\frac{S_{th}}{D} - {e}_{k, i})$
                    \State $\hat{E}.append(\hat{e}_{k, i})$
                \EndIf
            \EndFor
            \State $k = \arg \max \hat{E}$
            \State $\mathcal{B}.append(V_k)$
        \EndWhile
		\State Return $\mathcal{B}.remove(V_0)$
	\end{algorithmic}
	\label{alg:greedy}
	% \vspace{-20 pt}
\end{algorithm}

The time complexity of our greedy algorithm is $O(D \times N \log N)$, which reduces computation time by several orders of magnitude.

\textbf{Comparison of Brute-Force Method with Ours.} Here, we show the performance comparison of Brute-Force Method and our greedy algorithm in terms of PSNR and computation time. As the results in Table \ref{table:comparisonBFandgreedy_MipNeRF360}, \ref{table:comparisonBFandgreedy_llff} and \ref{table:comparisonBFandgreedy_real}, L-NeRF can achieve comparable PSNR
with Brute-Force method, however, the runtime of the Brute-Force method is several orders of magnitude larger than that of our proposed greedy algorithm, which makes the Brute-Force method impractical for incremental training scenarios. It should be noted that the runtime of the Brute-Force method increases exponentially with the size of the training data. For example, the `Bicycle' scene in Mip-NeRF360 contains 194 training images and `Horns' scene in LLFF contains 62 training images, but the Brute-Force method's runtime for these two scenes are 5 days 6 hours and 1 hour 47 min, respectively.

\begin{table}[!ht]
\centering
	\caption{Performance comparison of Brute-Force method and our IL-NeRF on the Mip-NeRF360. IL-NeRF can achieve comparable PSNR with Brute-Force method, however, the runtime of the Brute-Force method is several orders of magnitude larger than our proposed greedy algorithm.}
    \begin{tabular}{cccc}
    \toprule[1.5pt]
        Scene & Method & PSNR & running time\\
        \hline
        \multirow{2}{*}{Bicycle}& Brute-Force &  22.36 & 5 days 6 hours \\
        & Greedy (Ours) & 22.34 & 10.92 ms\\
        \hline
        \multirow{2}{*}{Bonsai}& Brute-Force & 28.96 & 10 days 8 hours \\
        & Greedy (Ours) & 28.96 & 25.57 ms\\
        \hline
        \multirow{2}{*}{Counter}& Brute-Force & 27.86 & 10 days 2 hours  \\
        & Greedy (Ours) & 27.82 & 23.78 ms\\ 
        \hline
        \multirow{2}{*}{Garden}& Brute-Force & 24.83 & 4 days 18 hours \\
        & Greedy (Ours) & 24.82 & 9.87 ms\\
        \hline
        \multirow{2}{*}{Kitchen}& Brute-Force & 29.34 & 11 days 6 hours\\
        & Greedy (Ours) & 29.34 & 28.39 ms\\
        \hline
        \multirow{2}{*}{Room}& Brute-Force & 31.49 & 12 days 10 hours\\
        & Greedy (Ours) & 31.45 & 37.58 ms\\
        \hline
        \multirow{2}{*}{Stump}& Brute-Force & 24.91 & 4 days 12 hours \\
        & Greedy (Ours) &  24.89 & 8.73 ms\\
        \toprule[1.5pt]
    \end{tabular}
\label{table:comparisonBFandgreedy_MipNeRF360}
\end{table}

\begin{table}[!ht]
\centering
	\caption{Performance comparison of Brute-Force method and our IL-NeRF on the LLFF. IL-NeRF can achieve comparable PSNR with Brute-Force method, however, the runtime of the Brute-Force method is several orders of magnitude larger than our proposed greedy algorithm.}
    \begin{tabular}{cccc}
    \toprule[1.5pt]
        Scene & Method & PSNR & running time\\
        \hline
        \multirow{2}{*}{Fern}& Brute-Force &  25.27 & 62.01 s \\
        & Greedy (Ours) & 25.26 & 4.38 ms\\
        \hline
        \multirow{2}{*}{Flower}& Brute-Force & 30.49 &  4.63 min\\
        & Greedy (Ours) & 30.49 & 6.81 ms\\
        \hline
        \multirow{2}{*}{Fortress}& Brute-Force & 29.45 & 14.17 min  \\
        & Greedy (Ours) & 29.45 & 8.78 ms\\ 
        \hline
        \multirow{2}{*}{Horns}& Brute-Force & 28.97 & 1 hour 47 min \\
        & Greedy (Ours) & 28.96 & 9.87 ms\\
        \hline
        \multirow{2}{*}{Leaves}& Brute-Force & 23.88 & 65.78 s\\
        & Greedy (Ours) & 23.88 & 4.95 ms\\
        \hline
        \multirow{2}{*}{Orchids}& Brute-Force & 23.67 & 57.84 s\\
        & Greedy (Ours) & 23.67 & 5.58 ms\\
        \hline
        \multirow{2}{*}{Room}& Brute-Force & 31.88 & 12.48 min \\
        & Greedy (Ours) &  31.88 & 8.73 ms\\
        \hline
        \multirow{2}{*}{Trex}& Brute-Force & 27.81 & 57.97 min \\
        & Greedy (Ours) &  27.81 & 9.37 ms\\
        \toprule[1.5pt]
    \end{tabular}
\label{table:comparisonBFandgreedy_llff}
\end{table}

\begin{table}[!ht]
\centering
	\caption{Performance comparison of Brute-Force method and our IL-NeRF on the NeRF-real360. IL-NeRF can achieve comparable PSNR with Brute-Force method, however, the runtime of the Brute-Force method is several orders of magnitude larger than our proposed greedy algorithm.}
    \begin{tabular}{cccc}
    \toprule[1.5pt]
        Scene & Method & PSNR & running time\\
        \hline
        \multirow{2}{*}{Pinecone}& Brute-Force &  22.96 & 4 days 10 hours \\
        & Greedy (Ours) & 22.93 & 9.58 ms\\
        \hline
        \multirow{2}{*}{Vasedeck}& Brute-Force & 26.24 & 5 days 2 hours \\
        & Greedy (Ours) & 26.15 & 10.61 ms\\
        \toprule[1.5pt]
    \end{tabular}
\label{table:comparisonBFandgreedy_real}
\end{table}

\subsection{Implementation Details}
We implement our framework following the architecture of Instant-NeRF \cite{muller2022instant,ngp-pl}. We use two separate Adam optimizers for NeRF and camera poses refinement respectively, with an initial learning rate of 0.01 for NeRF and an initial learning rate of 0.005 for pose refinement. The learning rate of the NeRF model decays every iteration by multiplying with 0.9954 (exponential decay), and the learning rate of the pose refinement decays every 100 iterations with a multiplier of 0.9. We train the network in each incremental step for Mip-NeRF360 with 30k iterations and $D=10$, LLFF with 5k iterations and $D=5$, and NeRF-real360 with 20k iterations and $D=10$. 

\section{More Comparisons for Results}
\label{sec:moreResults}
In the main text, we only show PSNR, SSIM and LPIPS for some scenes of the three datasets, and here we give the full results.
Table \ref{table:accuracyOnMipNeRF360} shows the results obtained by IL-NeRF and baseline methods on the Mip-NeRF360 dataset with seven real-world indoor and outdoor scenes. Similarly, Table \ref{table:accuracyOnLLFF} presents the results obtained on the LLFF dataset with eight forward-facing scenes. Additionally, Table \ref{table:accuracyOnNeRFReal360} also shows the results obtained on the NeRF-real360 dataset with two real-world object-orientation scenes. From the results, we can see that IL-NeRF outperforms the original NeRF and achieves comparable results with NeRF$^*$. 

\begin{table*}[!ht]
\centering
	\caption{Performance comparison on the Mip-NeRF360 dataset with the baselines: PSNR, SSIM, and LPIPS.  IL-NeRF outperforms the original NeRF, EWC, NeRF-SLAM and
achieves comparable results with NeRF$^*$.}
    \begin{tabular}{cc|c|cccc}
    \toprule[1.5pt]
        Scene & Method & Pose & \multicolumn{4}{c}{\textbf{PSNR $\Uparrow$ / SSIM $\Uparrow$ / LPIPS $\Downarrow$}}\\
        \cline{4-7}
        & & & $G^0$ & $G^1$ & $G^2$ & $G^3$\\
        \hline
        \multirow{4}{*}{Bicycle}& NeRF & Yes & 22.76 / 0.61 / 0.33 & 18.58 / 0.47 / 0.46 & 20.03 / 0.52 / 0.43 & 20.03 / 0.52 / 0.44 \\
        & EWC & Yes & 22.76 / 0.61 / 0.33 & 18.80 / 0.47 / 0.45 & 19.41 / 0.51 / 0.43 & 19.89 / 0.52 / 0.43 \\
        & NeRF$^*$ & Yes & 22.88 / 0.62 / 0.33 & \textbf{20.23} / \textbf{0.49} / \textbf{0.43} & 22.03 / 0.53 / \textbf{0.39} & 22.18 / 0.54 / 0.41 \\
        & NeRF-SLAM & No & 22.78 / 0.61 / 0.33 & 19.67 / 0.48 / 0.45 & 21.37 / 0.53 / 0.41 & 21.61 / 0.53 / 0.42 \\
        & IL-NeRF & No & \textbf{22.90} / \textbf{0.62} / \textbf{0.33}
        & 19.84 / 0.48 / 0.44 & \textbf{22.05} / \textbf{0.54} / 0.40 & \textbf{22.34} / \textbf{0.55} / \textbf{0.40} \\
        \hline
        \multirow{4}{*}{Bonsai}& NeRF & Yes & 33.30 / 0.93 / 0.07 & 25.47 / 0.75 / 0.25 & 23.53 / 0.66 / 0.34 & 22.12 / 0.68 / 0.35 \\
        & EWC & Yes & 33.30 / 0.93 / 0.07 & 25.62 / 0.75 / 0.25 & 22.35 / 0.66 / 0.33 & 21.51 / 0.68 / 0.34 \\
        & NeRF$^*$ & Yes & 33.48 / 0.93 / 0.07 & 29.93 / 0.88 / 0.15 & 28.03 / 0.84 / 0.18 & 28.18 / 0.84 / 0.21 \\
        & NeRF-SLAM & No & 33.32 / 0.93 / 0.07 & 29.13 / 0.84 / 0.21 & 28.01 / 0.79 / 0.29 & 26.85 / 0.80 / 0.29 \\
        & IL-NeRF & No & \textbf{33.54} / \textbf{0.93} / \textbf{0.07}
        & \textbf{30.73} / \textbf{0.89} / \textbf{0.12} & \textbf{29.77} / \textbf{0.86} / \textbf{0.16} & \textbf{28.96} / \textbf{0.85} / \textbf{0.18} \\
        \hline
        \multirow{4}{*}{Counter}& NeRF & Yes & 32.12 / 0.91 / 0.07 & 24.62 / 0.72 / 0.25 & 21.94 / 0.65 / 0.34 & 20.30 / 0.62 / 0.37 \\
        & EWC & Yes & 32.12 / 0.91 / 0.07 & 23.83 / 0.72 / 0.25 & 22.56 / 0.65 / 0.33 & 21.11 / 0.61 / 0.36 \\
        & NeRF$^*$ & Yes & \textbf{32.17} / \textbf{0.92} / 0.07 & 29.58 / 0.86 / 0.14 & 28.03 / 0.82 / 0.18 & \textbf{28.28} / \textbf{0.85} / 0.18 \\
        & NeRF-SLAM & No & 31.75 / 0.91 / 0.07 & 28.30 / 0.83 / 0.21 & 26.84 / 0.79 / 0.28 & 25.30 / 0.77 / 0.31 \\  
        & IL-NeRF & No & 32.13 / 0.91 / \textbf{0.07}
        & \textbf{29.63} / \textbf{0.87} / \textbf{0.12} & \textbf{28.56} / \textbf{0.85} / \textbf{0.15} & 27.82 / 0.83 / \textbf{0.17} \\ 
        \hline
        \multirow{4}{*}{Garden}& NeRF & Yes & 24.70 / 0.71 / 0.20
 & 22.34
 / 0.64 / 0.25 & 20.17 / 0.59 / 0.31 & 19.42 / 0.54 / 0.38 \\
        & EWC & Yes & 24.70 / 0.71 / 0.20 & 23.38 / 0.63 / 0.24 & 20.09 / 0.58 / 0.31 & 19.81 / 0.54 / 0.37 \\
        & NeRF$^*$ & Yes & 24.72 / 0.73 / 0.19 & \textbf{24.93} / \textbf{0.72} / \textbf{0.18} & 24.68 / 0.69 / 0.22 & 24.48 / 0.67 / \textbf{0.21} \\
        & NeRF-SLAM & No & 24.72 / 0.71 / 0.20 & 24.03 / 0.69 / 0.23 & 23.50 / 0.65 / 0.28 & 23.37 / 0.61 / 0.33 \\
        & IL-NeRF & No & \textbf{24.73} / \textbf{0.73} / \textbf{0.19}
        & 24.80 / 0.70 / 0.22 & \textbf{24.86} / \textbf{0.69} / \textbf{0.23} & \textbf{24.82} / \textbf{0.67} / 0.23 \\
        \hline
        \multirow{4}{*}{Kitchen}& NeRF & Yes & 31.17 / 0.91 / 0.08 & 27.01 / 0.75 / 0.25  & 21.42 / 0.70 / 0.31 & 23.69 / 0.75 / 0.24\\
        & EWC & Yes & 31.17 / 0.91 / 0.08 & 26.76 / 0.74 / 0.25 & 22.09 / 0.70 / 0.31 & 23.39 / 0.74 / 0.23 \\
        & NeRF$^*$ & Yes & 31.05 / 0.91 / 0.07 & 29.72 / 0.88 / 0.13 & 29.33 / 0.85 / 0.15 & 29.18 / 0.84 / 0.14 \\
        & NeRF-SLAM & No & 30.87 / 0.90 / 0.09 & 29.63 / 0.85 / 0.20 & 27.65 / 0.81 / 0.24 & 27.71 / 0.82 / 0.20 \\
        & IL-NeRF & No & \textbf{31.27} / \textbf{0.92} / \textbf{0.07}
        & \textbf{30.66} / \textbf{0.89} / \textbf{0.10} & \textbf{29.84} / \textbf{0.87} / \textbf{0.12} & \textbf{29.34} / \textbf{0.86} / \textbf{0.13} \\
        \hline
        \multirow{4}{*}{Room}& NeRF & Yes & 35.98 / 0.96 / 0.04 & 30.78 / 0.91 / 0.09 & 26.34 / 0.80 / 0.21 & 27.44 / 0.86 / 0.16 \\
        & EWC & Yes & 35.98 / 0.96 / 0.04 & 31.84 / 0.90 / 0.09 & 27.38 / 0.79 / 0.20 & 28.08 / 0.86 / 0.16 \\
        & NeRF$^*$ & Yes & \textbf{36.18} / 0.96 / \textbf{0.03} & 33.93 / \textbf{0.95} / 0.05 & 32.03 / 0.92 / 0.08 & \textbf{31.99} / \textbf{0.93} / \textbf{0.06} \\
        & NeRF-SLAM & No & 35.74 / 0.94 / 0.08 & 33.20 / 0.93 / 0.07 & 30.36 / 0.88 / 0.17 & 30.73 / 0.89 / 0.13 \\
        & IL-NeRF & No & 36.04 / \textbf{0.96} / 0.04
        & \textbf{34.02} / 0.94 / \textbf{0.04} & \textbf{32.35} / \textbf{0.92} / \textbf{0.07} & 31.45 / 0.91 / 0.09 \\
        \hline
        \multirow{4}{*}{Stump}& NeRF & Yes & 25.62 / 0.77 / 0.28 & 22.30 / 0.52 / 0.38 & 21.25 / 0.46 / 0.42 & 20.55 / 0.44 / 0.47 \\
        & EWC & Yes & 25.62 / 0.77 / 0.28 & 22.55 / 0.51 / 0.37 & 21.09 / 0.45 / 0.42 & 21.48 / 0.44 / 0.46 \\
        & NeRF$^*$ & Yes &  \textbf{26.18} / \textbf{0.79} / \textbf{0.27} & \textbf{25.93} / 0.64 / 0.37 & \textbf{25.12} / \textbf{0.62} / 0.38 & \textbf{25.18} / \textbf{0.64} / 0.39 \\
        & NeRF-SLAM & No & 24.98 / 0.74 / 0.31 & 24.76 / 0.61 / 0.36 & 24.05 / 0.56 / 0.40 & 23.93 / 0.57 / 0.43 \\
        & IL-NeRF & No &25.96 / 0.77 / 0.28
        & 25.75 / \textbf{0.66} / \textbf{0.32} & 25.09 / 0.60 / \textbf{0.37} & 24.89 / 0.58 / \textbf{0.37} \\
        \toprule[1.5pt]
    \end{tabular}
\label{table:accuracyOnMipNeRF360}
\end{table*}

\begin{table*}[!ht]
\centering
	\caption{Performance comparison on the LLFF dataset with the baselines: PSNR, SSIM, and LPIPS. IL-NeRF outperforms the original NeRF, EWC, NeRF-SLAM and
achieves comparable results with NeRF$^*$.}
    \begin{tabular}{cc|c|cccc}
    \toprule[1.5pt]
        Scene & Method & Pose & \multicolumn{4}{c}{\textbf{PSNR $\Uparrow$ / SSIM $\Uparrow$ / LPIPS $\Downarrow$}}\\
        \cline{4-7}
        & & & $G^0$ & $G^1$ & $G^2$ & $G^3$\\
        \hline
        \multirow{4}{*}{Fern}& NeRF & Yes & 29.19 / 0.90 / 0.06 & 24.58 / 0.80 / 0.19 & 23.21 / 0.67 / 0.24 & 22.43 / 0.65 / 0.25 \\
        & EWC & Yes & 29.19 / 0.90 / 0.06 &  24.88 / 0.79 / 0.19 & 22.60 / 0.66 / 0.23 & 23.32 / 0.64 / 0.25\\
        & NeRF$^*$ & Yes &  29.26 / 0.90 / 0.06 & \textbf{26.71} / \textbf{0.88} / \textbf{0.09} & \textbf{25.79} / \textbf{0.85} / \textbf{0.12} & 25.19 / \textbf{0.82} / \textbf{0.13}\\
        & NeRF-SLAM & No & 28.77 / 0.88 / 0.16 &  25.22 / 0.85 / 0.15 & 24.95 / 0.79 / 0.23 & 24.33 / 0.77 / 0.21\\
        & IL-NeRF & No &  \textbf{29.30} / \textbf{0.90} / \textbf{0.06} & 26.68 / 0.87 / 0.10 & 25.63 / 0.81 / 0.13 & \textbf{25.26} / 0.80 / 0.15 \\
        \hline
        \multirow{4}{*}{Flower}& NeRF & Yes & 34.12 / 0.96 / 0.01 & 30.30 / 0.91 / 0.02 & 28.40 / 0.90 / 0.03 & 27.50 / 0.88/ 0.04  \\
        & EWC & Yes &  34.12 / 0.96 / 0.01 & 29.84 / 0.90 / 0.02 & 28.16 / 0.90 / 0.03 & 27.14 / 0.88 / 0.03\\
        & NeRF$^*$ & Yes & \textbf{34.28} / 0.96 / 0.01 & 31.76 / 0.93 / 0.01 & 30.98 / 0.93 / 0.02 & \textbf{30.68} / 0.93 / 0.02\\
        & NeRF-SLAM & No &  33.28 / 0.96 / 0.01 & 31.34 / 0.92 / 0.01 & 30.29 / 0.92 / 0.02 & 29.72 / 0.91 / 0.03\\
        & IL-NeRF & No & 34.22 / \textbf{0.96} / \textbf{0.01} & \textbf{31.81} / \textbf{0.94} / \textbf{0.01} & \textbf{31.11} / \textbf{0.94} / \textbf{0.02} & 30.49 / \textbf{0.93} / \textbf{0.02}  \\
        \hline
        \multirow{4}{*}{Fortress}& NeRF & Yes & 31.56 / 0.85 / 0.11 & 29.38 / 0.80 / 0.15 & 27.05 / 0.78 / 0.17 & 25.39 / 0.78 / 0.16\\
        & EWC & Yes & 31.56 / 0.85 / 0.11 & 29.41 / 0.79 / 0.15 & 25.83 / 0.77 / 0.16 & 24.40 / 0.78 / 0.15\\
        & NeRF$^*$ & Yes & \textbf{31.75} / \textbf{0.86} / 0.11 & \textbf{31.83} / 0.84 / \textbf{0.09} & \textbf{30.90} / \textbf{0.86} / 0.14 & \textbf{29.81} / \textbf{0.85} / \textbf{0.11} \\
        & NeRf-SLAM & No & 31.08 / 0.82 / 0.12 & 31.09 / 0.82 / 0.12 & 29.77 / 0.83 / 0.15 & 28.53 / 0.82 / 0.14\\
        & IL-NeRF & No & 31.69 / 0.85 / \textbf{0.11} & 31.02 / \textbf{0.84} / 0.10 & 30.33 / 0.84 / \textbf{0.11} & 29.45 / 0.83 / 0.12\\ 
        \hline
        \multirow{4}{*}{Horns}& NeRF & Yes & 29.78 / 0.86 / 0.09 & 27.04 / 0.75 / 0.09 & 26.04 / 0.74 / 0.12 & 24.01 / 0.70 / 0.14\\
        & EWC & Yes & 29.78 / 0.86 / 0.09 & 26.77 / 0.75 / 0.09 & 27.08 / 0.74 / 0.11 &  24.68 / 0.69 / 0.14 \\
         & NeRF$^*$ & Yes & 29.86 / 0.89 / 0.07 & \textbf{29.67} / 0.89 / \textbf{0.06} & \textbf{29.24} / 0.89 / 0.07 & 28.87 / 0.87 / \textbf{0.08}\\
        & NeRF-SLAM & No & 28.86 / 0.83 / 0.10 & 28.77 / 0.84 / 0.08 & 28.19 / 0.83 / 0.10 &  27.56 / 0.82 / 0.12 \\
        & IL-NeRF & No & \textbf{29.92} / \textbf{0.89} / \textbf{0.07} & 29.50 / \textbf{0.89} / 0.07 & 29.01 / \textbf{0.89} / \textbf{0.07} & \textbf{28.96} / \textbf{0.87} / 0.09\\
        \hline
        \multirow{4}{*}{Leaves}& NeRF & Yes & 25.51 / 0.90 / 0.06 & 22.12 / 0.79 / 0.13 & 21.03 / 0.75 / 0.15 & 20.62 / 0.73 / 0.16\\
        & EWC & Yes &  25.51 / 0.90 / 0.06 & 21.18 / 0.79 / 0.13 & 21.96 / 0.74 / 0.15 & 20.39 / 0.72 / 0.16\\
        & NeRF$^*$ & Yes & 25.58 / 0.90 / 0.06 & \textbf{24.81} / \textbf{0.89} / 0.07 & 24.23 / 0.87 / 0.07 & 23.84 / 0.86 / 0.08\\
        & NeRF-SLAM & No &  24.83 / 0.87 / 0.08 & 23.93 / 0.86 / 0.10 & 23.27 / 0.83 / 0.12 & 22.86 / 0.82 / 0.13\\
        & IL-NeRF & No & \textbf{25.62} / \textbf{0.90} / \textbf{0.06} & 24.74 / 0.88 / \textbf{0.07} & \textbf{24.26} / \textbf{0.87} / \textbf{0.07} & \textbf{23.88} / \textbf{0.86} / \textbf{0.08}\\
        \hline
        \multirow{4}{*}{Orchids}& NeRF & Yes &  25.68 / 0.85 / 0.08 & 22.78 / 0.77 / 0.10 & 21.43 / 0.74 / 0.12 & 20.77 / 0.71 / 0.14\\
        & EWC & Yes &  25.68 / 0.85 / 0.08 & 22.18 / 0.77/ 0.10 & 21.41 / 0.73 / 0.12 & 20.23 / 0.70 / 0.14 \\
        & NeRF$^*$ & Yes &  \textbf{25.88} / \textbf{0.87} / 0.08 & \textbf{24.37} / \textbf{0.84} / 0.10 & 23.69 / 0.80 / 0.12 & 23.59 / \textbf{0.79} / \textbf{0.12}\\
        & NeRF-SLAM & No &  24.32 / 0.81 / 0.09 & 23.94 / 0.82/ 0.10 & 23.26 / 0.78 / 0.12 & 22.76 / 0.76 / 0.13 \\
        & IL-NeRF & No & 25.78 / 0.86 / \textbf{0.08} & 24.17 / 0.82 / \textbf{0.10} & \textbf{23.89} / \textbf{0.80} / \textbf{0.12} & \textbf{23.67} / 0.77 / 0.13 \\
        \hline
        \multirow{4}{*}{Room}& NeRF & Yes & 32.35 / 0.92 / 0.09 & 29.58 / 0.89 / 0.10 & 28.82 / 0.89 / 0.11 &  30.59 / 0.92 / 0.07\\
        & EWC & Yes & 31.14 / 0.92 / 0.09 & 28.46 / 0.88 / 0.10 & 28.78 / 0.89 / 0.11 & 30.29 / 0.91 / 0.08 \\
        & NeRF$^*$ & Yes &  32.26 / 0.92 / 0.09 & 31.52 / 0.92 / 0.08 & 31.21 / 0.92 / 0.08 & \textbf{31.98} / 0.92 / 0.08\\
        & NeRF-SLAM & No & 30.84 / 0.89 / 0.10 & 31.02 / 0.91 / 0.09 & 30.76 / 0.91 / 0.10 & 31.63 / 0.92 / 0.07 \\
        & IL-NeRF & No & \textbf{32.50} / \textbf{0.92} / \textbf{0.08} & \textbf{31.76} / \textbf{0.92} / \textbf{0.08} & \textbf{31.58} / \textbf{0.92} / \textbf{0.08} & 31.88 / \textbf{0.92} / \textbf{0.07}\\
        \hline
        \multirow{4}{*}{Trex}& NeRF & Yes &  28.50 / 0.90 / 0.07 & 27.29 / 0.89 / 0.07 & 26.40 / 0.88 / 0.07 & 26.24 / 0.86 / 0.09\\
        & EWC & Yes &  28.50 / 0.90 / 0.07 & 26.42 / 0.88 / 0.07 & 25.97 / 0.89 / 0.07 & 25.18 / 0.91 / 0.09 \\ 
        & NeRF$^*$ & Yes & \textbf{28.74} / \textbf{0.91} / \textbf{0.06} & \textbf{28.32} / 0.90 / 0.06 & \textbf{28.11} / 0.90 / 0.07 & \textbf{27.98} / 0.90 / 0.06 \\
        & NeRF-SLAM & No &  27.26 / 0.90 / 0.07 & 28.05 / 0.89 / 0.06 & 27.60 / 0.89 / 0.07 & 27.37 / 0.88 / 0.08 \\ 
        & IL-NeRF & No & 28.70 / 0.90 / 0.07 & 28.14 / \textbf{0.90} / \textbf{0.06} & 27.90 / \textbf{0.90} / \textbf{0.07} & 27.81 / \textbf{0.90} / \textbf{0.06}\\
        \toprule[1.5pt]
    \end{tabular}
\label{table:accuracyOnLLFF}
\end{table*}

\begin{table*}[!ht]
\centering
	\caption{Performance comparison on the NeRF-real360 dataset with the baselines: PSNR, SSIM, and LPIPS. IL-NeRF outperforms the original NeRF, EWC, NeRF-SLAM and
achieves comparable results with NeRF$^*$.}
    \begin{tabular}{cc|c|cccc}
    \toprule[1.5pt]
        Scene & Method & Pose & \multicolumn{4}{c}{\textbf{PSNR $\Uparrow$ / SSIM $\Uparrow$ / LPIPS $\Downarrow$}}\\
        \cline{4-7}
        & & & $G^0$ & $G^1$ & $G^2$ & $G^3$\\
        \hline
        \multirow{4}{*}{Pinecone}& NeRF & Yes & 26.22 / 0.84 / 0.16 & 22.90 / 0.64 / 0.24 & 21.15 / 0.58 / 0.33 & 18.94 / 0.49 / 0.41 \\
        & EWC & Yes & 26.22 / 0.84 / 0.16 & 22.70 / 0.63 / 0.24 & 21.42 / 0.57 / 0.32 & 18.81 / 0.48 / 0.41 \\
        & NeRF$^*$ & Yes & \textbf{26.88} / \textbf{0.89} / 0.12 & 24.23 / \textbf{0.79} / \textbf{0.16} & \textbf{24.03} / 0.73 / \textbf{0.19} & \textbf{23.18} / \textbf{0.74} / \textbf{0.21} \\
        & NeRF-SLAM & No & 25.63 / 0.81 / 0.18 & 24.09 / 0.73 / 0.22 & 23.01 / 0.68 / 0.29 & 21.79 / 0.65 / 0.34 \\
        & IL-NeRF & No & 26.31 / 0.87 / \textbf{0.10}
        & \textbf{24.56} / 0.78 / 0.17 & 23.78 / \textbf{0.74} / 0.20 & 22.93 / 0.72 / 0.23 \\
        \hline
        \multirow{4}{*}{Vasedeck}& NeRF & Yes & 29.03 / 0.85 / 0.07 & 23.99 / 0.70 / 0.26 & 22.73 / 0.69 / 0.24 & 21.57 / 0.64 / 0.31 \\
        & EWC & Yes & 29.03 / 0.85 / 0.07 & 24.36 / 0.69 / 0.25 & 22.25 / 0.68 / 0.24 & 20.52 / 0.64 / 0.30 \\
        & NeRF$^*$ & Yes & 29.27 / 0.86 / 0.07 & \textbf{27.93} / \textbf{0.85} / 0.12 & 26.03 / 0.74 / 0.16 & \textbf{26.18} / 0.74 / 0.18 \\
        & NeRF-SLAM & No & 27.98 / 0.79 / 0.11 & 26.41 / 0.77 / 0.21 & 25.10 / 0.72 / 0.21 & 24.62 / 0.71 / 0.26 \\
        & IL-NeRF & No & \textbf{29.48} / \textbf{0.86} / \textbf{0.07}
        & 27.38 / 0.82 / \textbf{0.10} & \textbf{26.11} / \textbf{0.76} / \textbf{0.14} & 26.15 / \textbf{0.75} / \textbf{0.17} \\
        \toprule[1.5pt]
    \end{tabular}
\label{table:accuracyOnNeRFReal360}
\end{table*}

Furthermore, we compare the original NeRF and IL-NeRF on two scenes, the ’Kitchen’ and `Counter' scenes in the Mip-NeRF360 dataset. Here, we give more visualization results of original NeRF and IL-NeRF on all scenes of three datasets.

\begin{figure*}[tt]
	\centering
	\includegraphics[width=\linewidth]{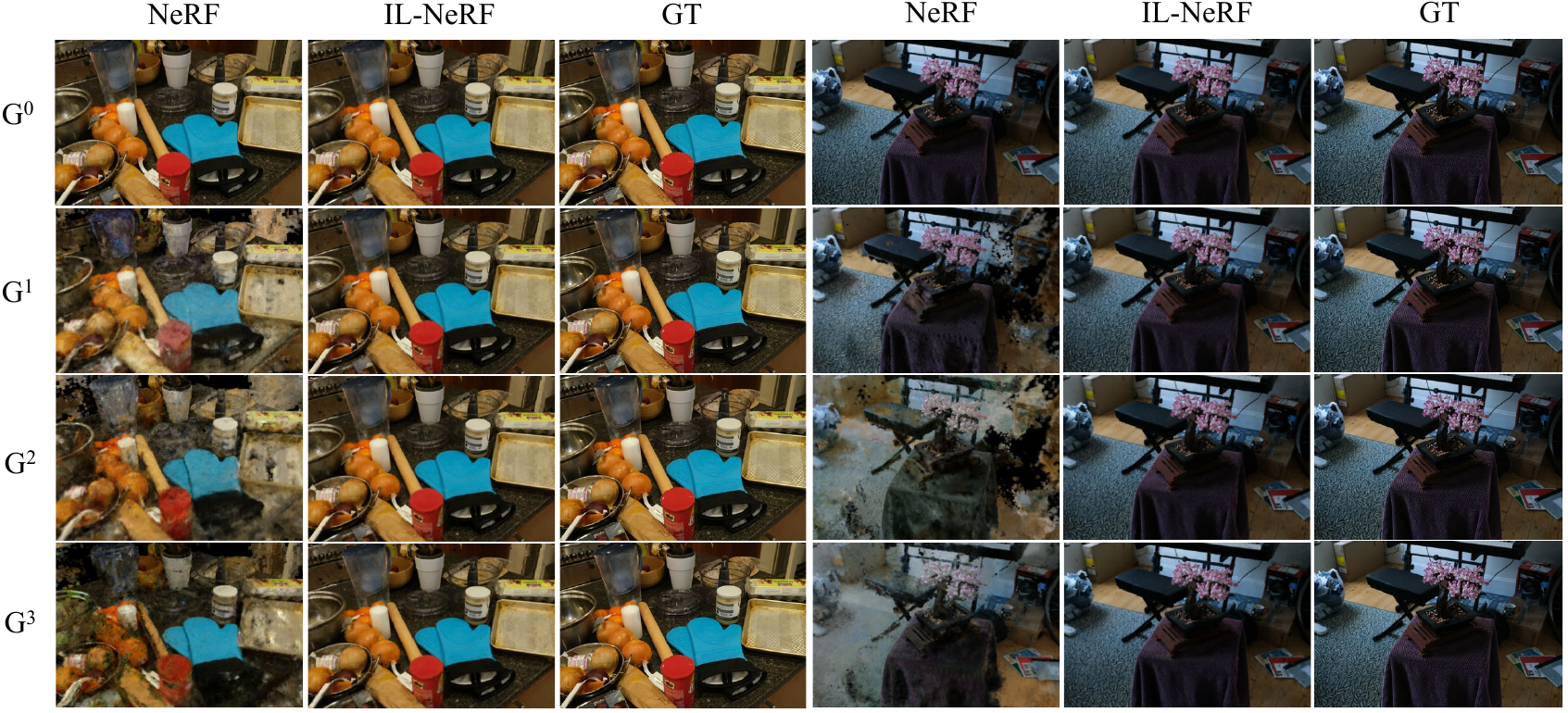}
	\caption{Qualitative comparison of the original NeRF and IL-NeRF on the rendering images in the first image data after each incremental training. GT means the ground truth of the training image. The original NeRF demonstrates severe catastrophic forgetting, leading to the loss of early-task scene information. In contrast, IL-NeRF is able to preserve the scene of interest throughout the training process. Testsets are the scenes 'Counter' and 'Bonsai' in the Mip-NeRF36 dataset.}
	\label{fig:supplement_counter_bonsai}
	%	\vspace{-10 pt}
\end{figure*}

\begin{figure*}[tt]
	\centering
	\includegraphics[width=\linewidth]{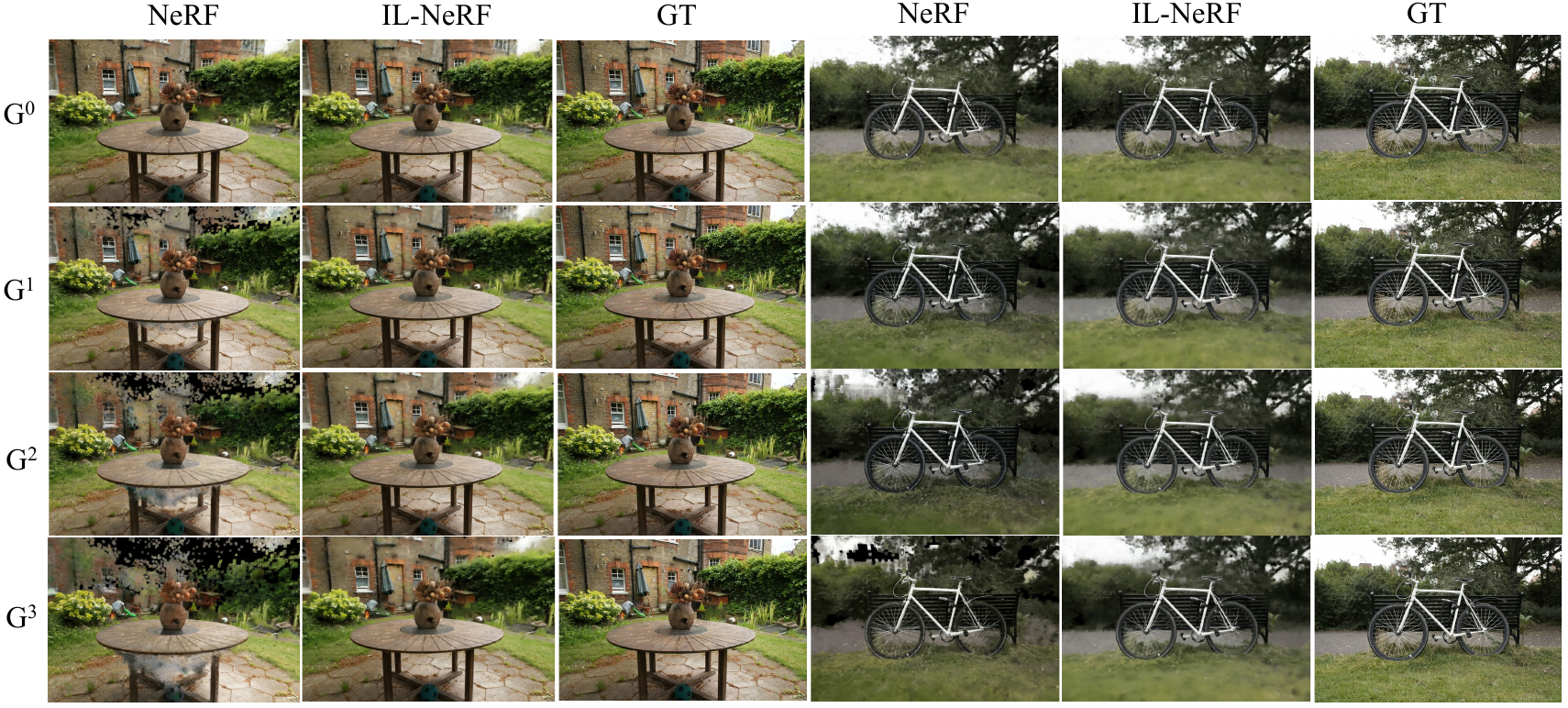}
	\caption{Qualitative comparison of the original NeRF and IL-NeRF on the rendering images in the first image data after each incremental training. GT means the ground truth of the training image. The original NeRF demonstrates severe catastrophic forgetting, leading to the loss of early-task scene information. In contrast, IL-NeRF is able to preserve the scene of interest throughout the training process. Testsets are the scenes 'Garden' and 'Bicycle' in the Mip-NeRF36 dataset.}
	\label{fig:supplement_garden_bicycle}
	%	\vspace{-10 pt}
\end{figure*}

\begin{figure*}[tt]
	\centering
	\includegraphics[width=\linewidth]{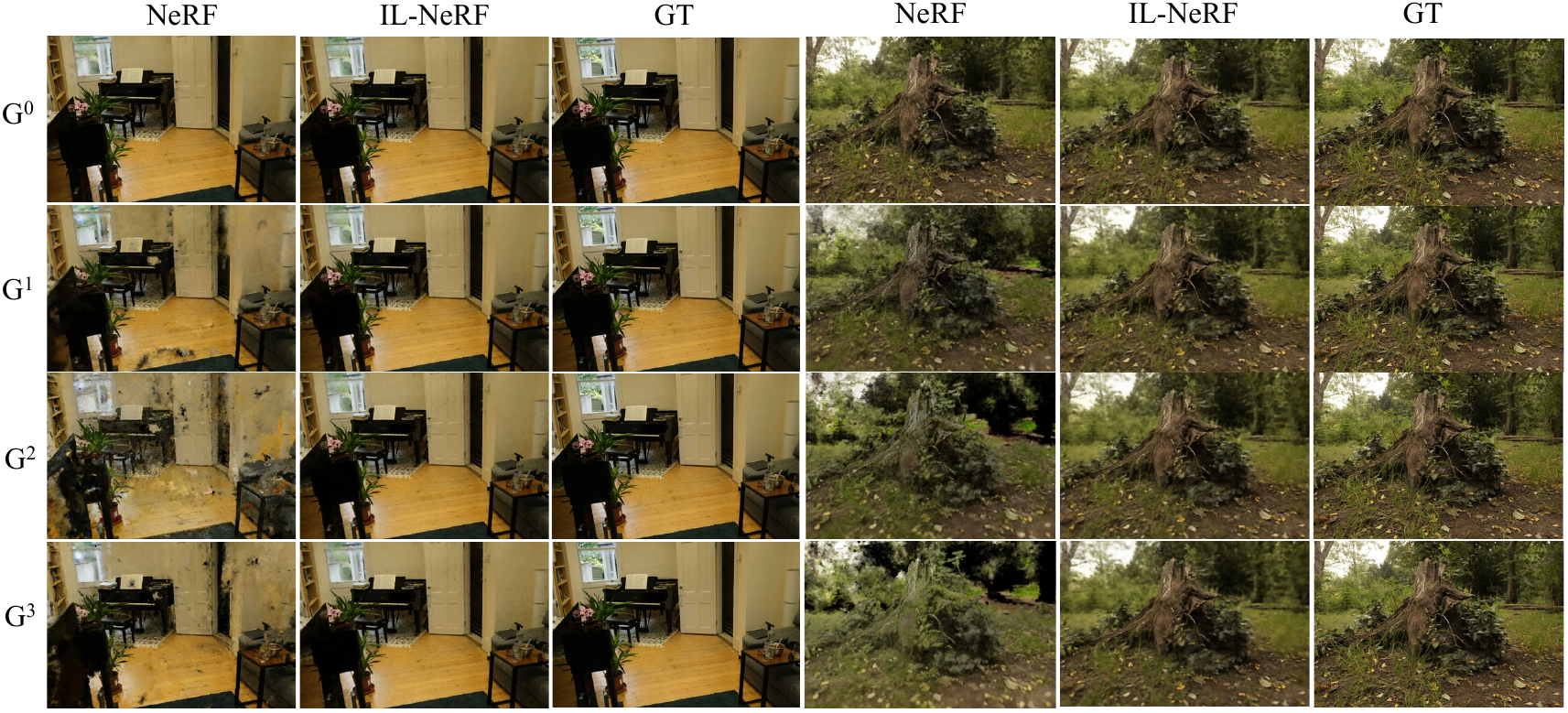}
	\caption{Qualitative comparison of the original NeRF and IL-NeRF on the rendering images in the first image data after each incremental training. GT means the ground truth of the training image. The original NeRF demonstrates severe catastrophic forgetting, leading to the loss of early-task scene information. In contrast, IL-NeRF is able to preserve the scene of interest throughout the training process. Testsets are the scenes 'Room' and 'Stump' in the Mip-NeRF36 dataset.}
	\label{fig:supplement_room_stump}
	%	\vspace{-10 pt}
\end{figure*}

\begin{figure*}[tt]
	\centering
	\includegraphics[width=\linewidth]{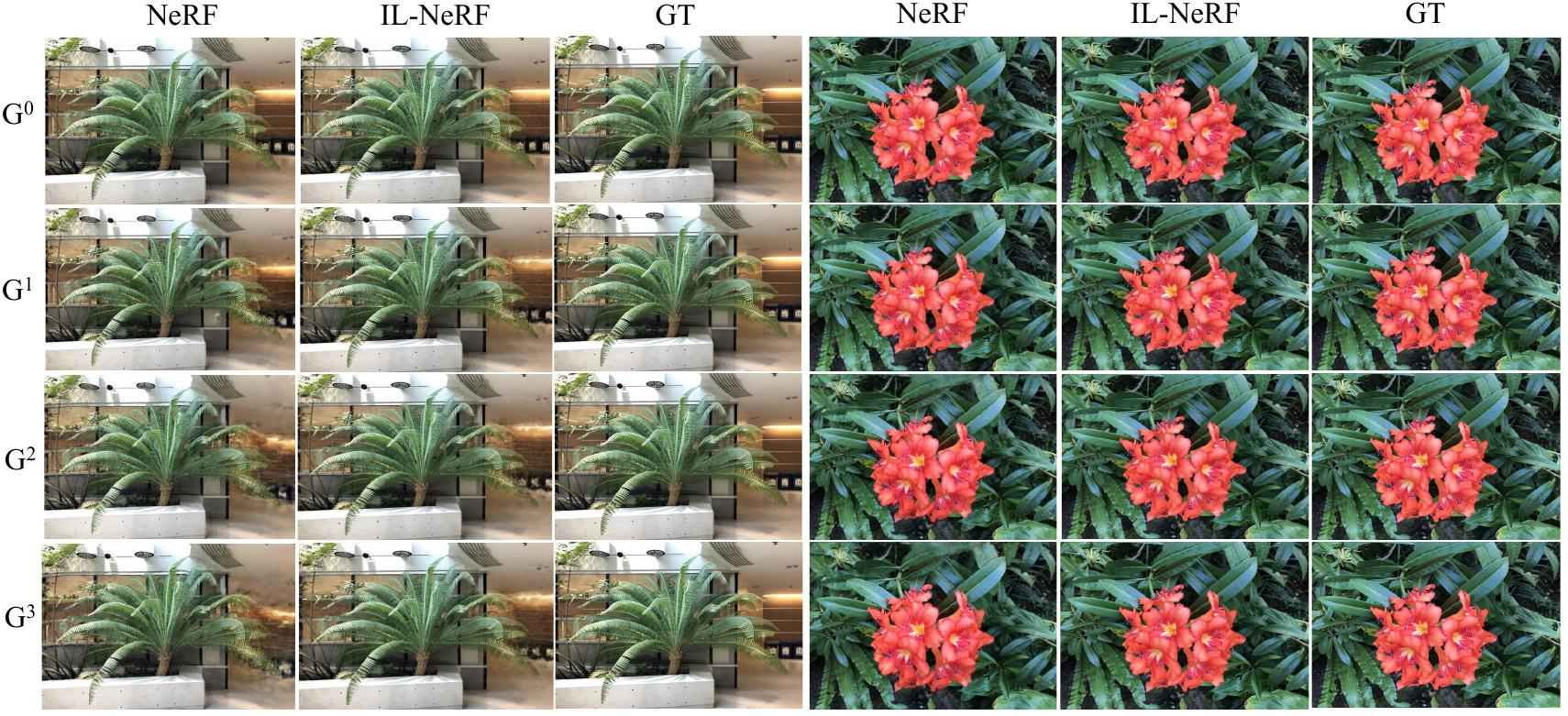}
	\caption{Qualitative comparison of the original NeRF and IL-NeRF on the rendering images in the first image data after each incremental training. GT means the ground truth of the training image. The original NeRF demonstrates severe catastrophic forgetting, leading to the loss of early-task scene information. In contrast, IL-NeRF is able to preserve the scene of interest throughout the training process. Testsets are the scenes 'Fern' and 'Flower' in the LLFF dataset.}
	\label{fig:supplement_fern_flower}
	%	\vspace{-10 pt}
\end{figure*}

\begin{figure*}[tt]
	\centering
	\includegraphics[width=\linewidth]{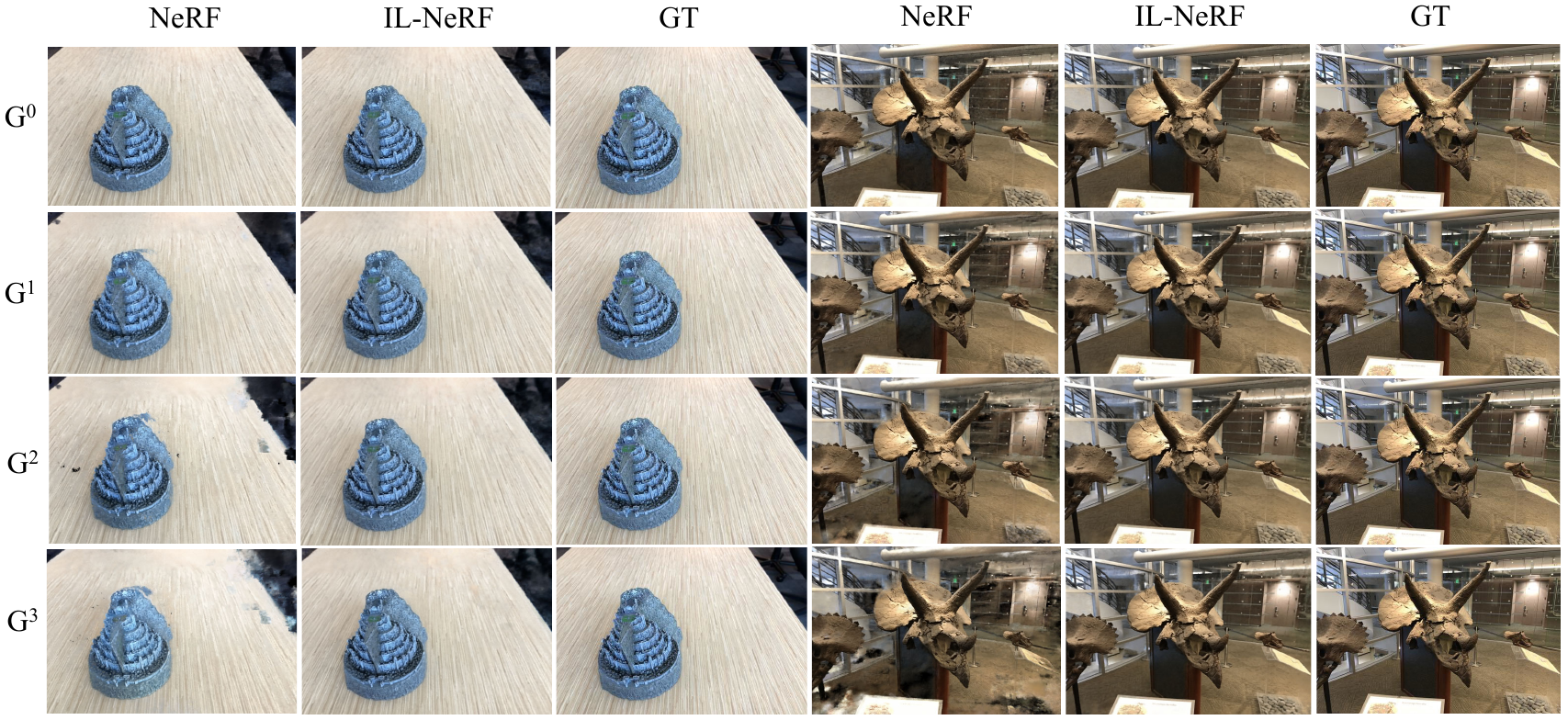}
	\caption{Qualitative comparison of the original NeRF and IL-NeRF on the rendering images in the first image data after each incremental training. GT means the ground truth of the training image. The original NeRF demonstrates severe catastrophic forgetting, leading to the loss of early-task scene information. In contrast, IL-NeRF is able to preserve the scene of interest throughout the training process. Testsets are the scenes 'Fortress' and 'Horns' in the LLFF dataset.}
	\label{fig:supplement_fortress_horns}
	%	\vspace{-10 pt}
\end{figure*}

\begin{figure*}[tt]
	\centering
	\includegraphics[width=\linewidth]{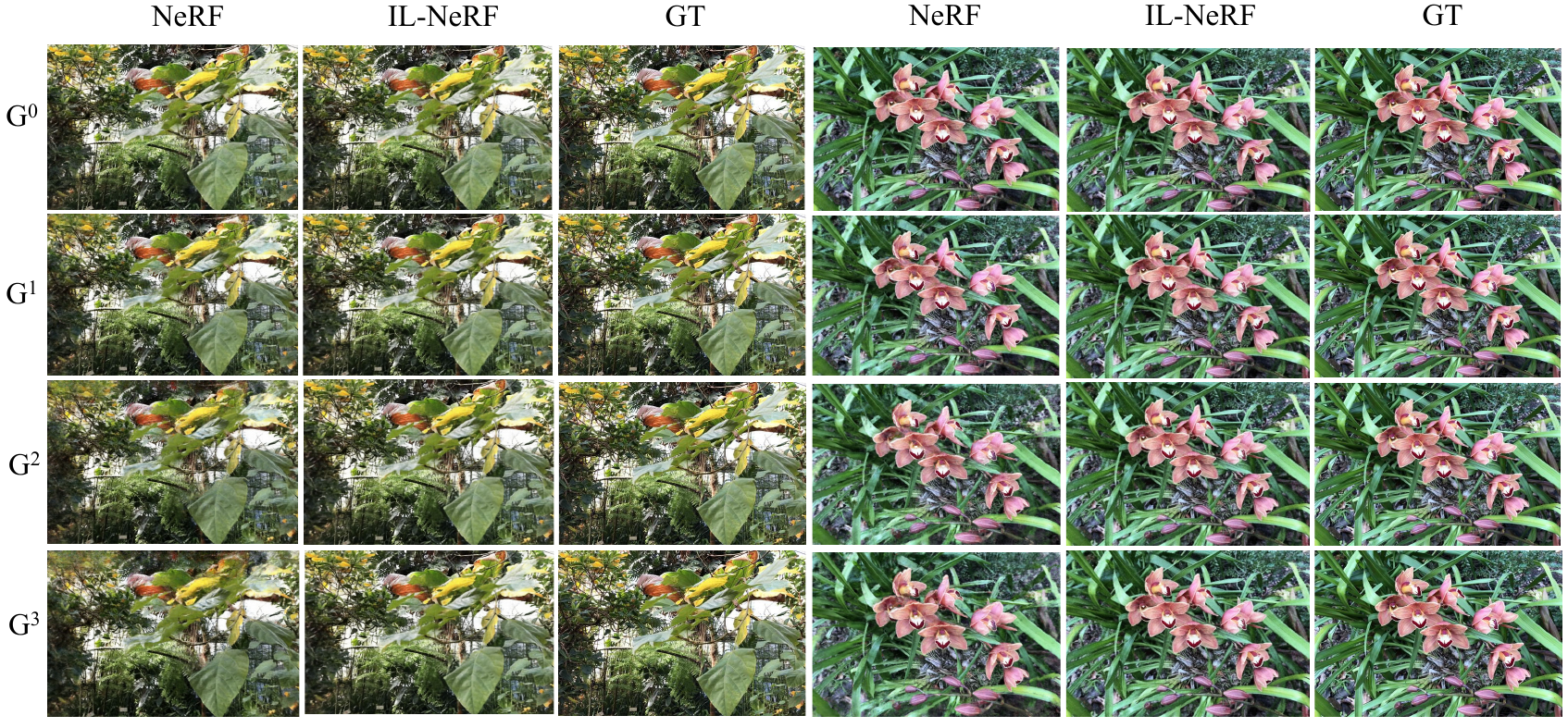}
	\caption{Qualitative comparison of the original NeRF and IL-NeRF on the rendering images in the first image data after each incremental training. GT means the ground truth of the training image. The original NeRF demonstrates severe catastrophic forgetting, leading to the loss of early-task scene information. In contrast, IL-NeRF is able to preserve the scene of interest throughout the training process. Testsets are the scenes 'Leaves' and 'Orchids' in the LLFF dataset.}
	\label{fig:supplement_leaves_orchids}
	%	\vspace{-10 pt}
\end{figure*}

\begin{figure*}[tt]
	\centering
	\includegraphics[width=\linewidth]{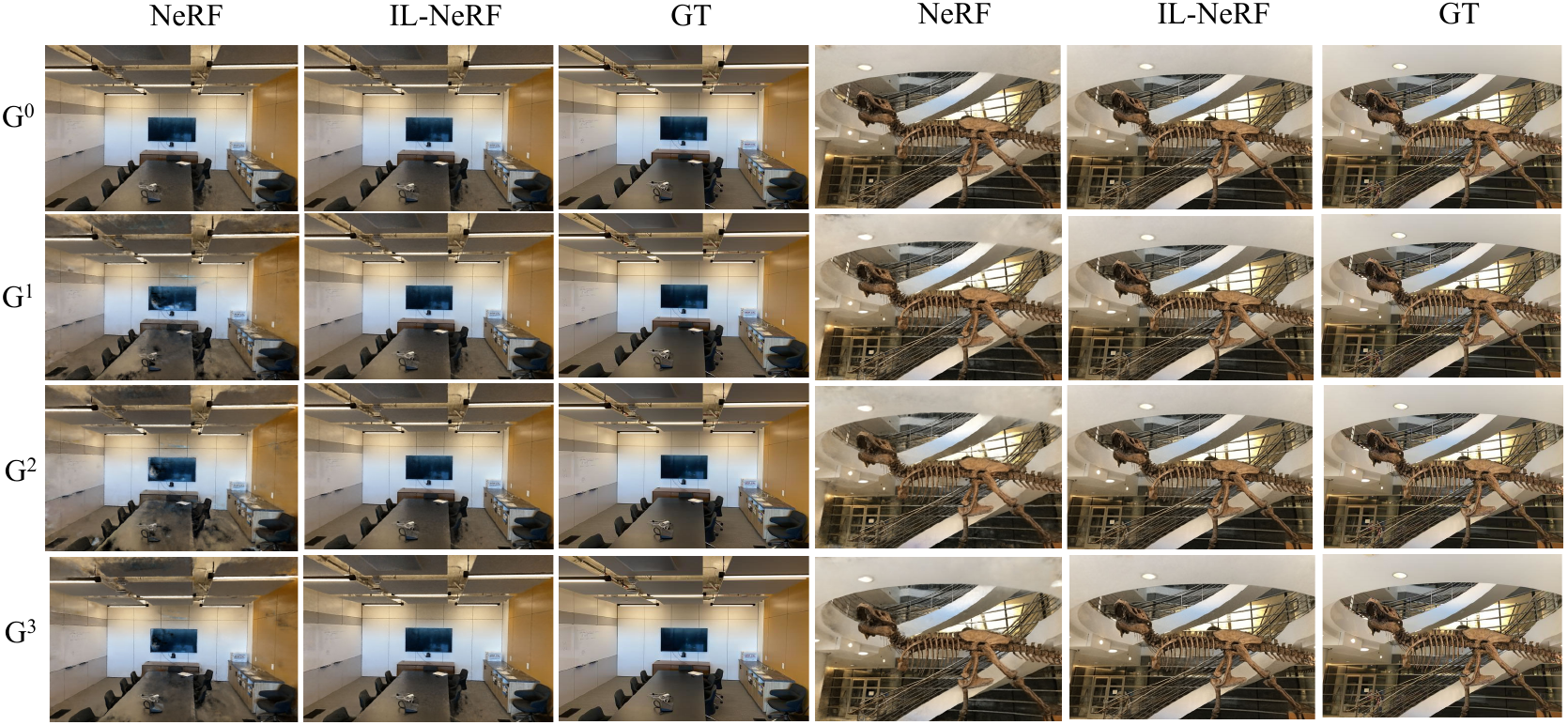}
	\caption{Qualitative comparison of the original NeRF and IL-NeRF on the rendering images in the first image data after each incremental training. GT means the ground truth of the training image. The original NeRF demonstrates severe catastrophic forgetting, leading to the loss of early-task scene information. In contrast, IL-NeRF is able to preserve the scene of interest throughout the training process. Testsets are the scenes 'Room' and 'Trex' in the LLFF dataset.}
	\label{fig:supplement_room_trex}
	%	\vspace{-10 pt}
\end{figure*}

\begin{figure*}[tt]
	\centering
	\includegraphics[width=\linewidth]{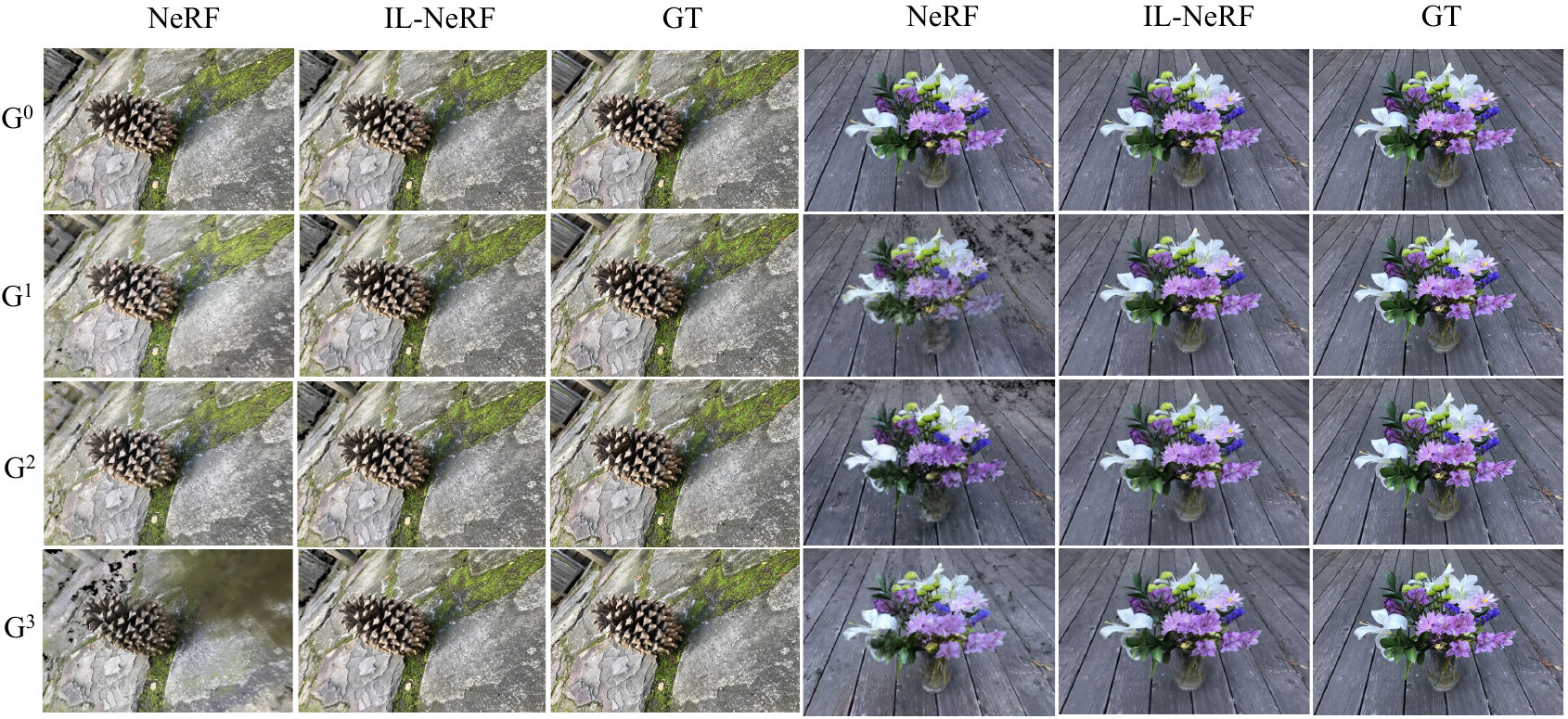}
	\caption{Qualitative comparison of the original NeRF and IL-NeRF on the rendering images in the first image data after each incremental training. GT means the ground truth of the training image. The original NeRF demonstrates severe catastrophic forgetting, leading to the loss of early-task scene information. In contrast, IL-NeRF is able to preserve the scene of interest throughout the training process. Testset is the scenes 'Pinecone' and Vasedeskin the NeRF-real360 dataset.}
	\label{fig:supplement_pinecone}
	%	\vspace{-10 pt}
\end{figure*}

Figure \ref{fig:supplement_counter_bonsai} to Figure \ref{fig:supplement_pinecone} provide additional insight by presenting a
qualitative comparison of the performance of the original NeRF and IL-NeRF. Specifically, we demonstrate the rendering results on the first task after each incremental training.
It is evident that the original NeRF suffers from the catastrophic forgetting problem, resulting in images with significant distortions such as noise and blur, whereas IL-NeRF generates highly realistic images with quality comparable to
the ground truth. This observation indicates that IL-NeRF is highly effective in mitigating the forgetting problem and addressing the coordinate shifting issue. 

\textbf{Video Demo.} To further show the performance of IL-NeRF, we post a video demonstration in the supplementary material, named `sm$\_$video.mp4'. In this video, we show rendered images from all baselines and IL-NeRF.

\section{Ablation Study} \label{sec:ablation}
In the main paper, we analyze the effectiveness of the camera coordinate alignment and the pose refinement that has been added to IL-NeRF on the scene 'Garden'. Here, we give more numerical results for the ablation study.

Table \ref{table:supplement_accuracyOnMipNeRF360} to Table \ref{table:supplement_accuracyOnNeRFReal360} shows the performance comparison of the original NeRF, NeRF$^*$, IL-NeRF, IL-NeRF w/o TM, and IL-NeRF w/o PR on all three datasets. As we can see, IL-NeRF outperforms the original NeRF and achieves comparable results with NeRF$^*$. The results reveal a significant decline in performance on all test data without the transfer matrices (i.e., IL-NeRF w/o TM). This decline can be attributed to separate camera pose estimation for two tasks resulting in camera poses in two independent coordinate systems, which could
mislead the model during training, leading to decreased performance. The results of IL-NeRF w/o PR, indicate that IL-NeRF with pose refinement outperforms IL-NeRF without it as the aligned  camera poses by the transfer matrices may still contain noise and inaccuracies.

\begin{figure*}[tt]
    \centering
    \begin{subfigure}{0.23\linewidth}
    \includegraphics[width=\linewidth]{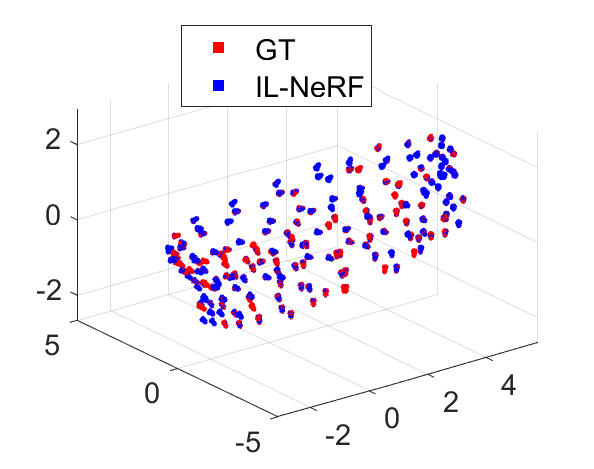}
    \caption{Bicycle}
    \end{subfigure}
    \begin{subfigure}{0.23\linewidth}
    \includegraphics[width=\linewidth]{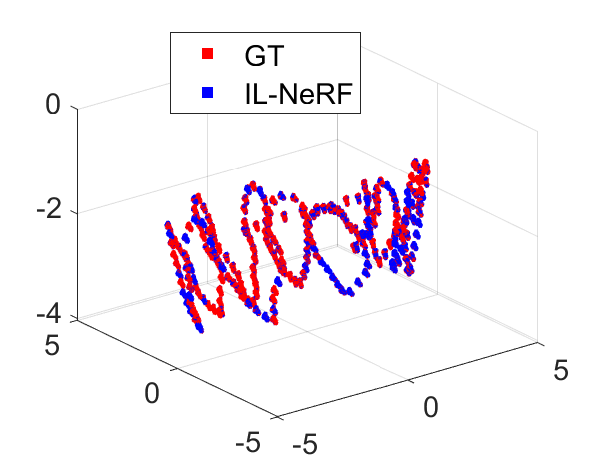}
    \caption{Counter}
    \end{subfigure}
    \begin{subfigure}{0.23\linewidth}
    \includegraphics[width=\linewidth]{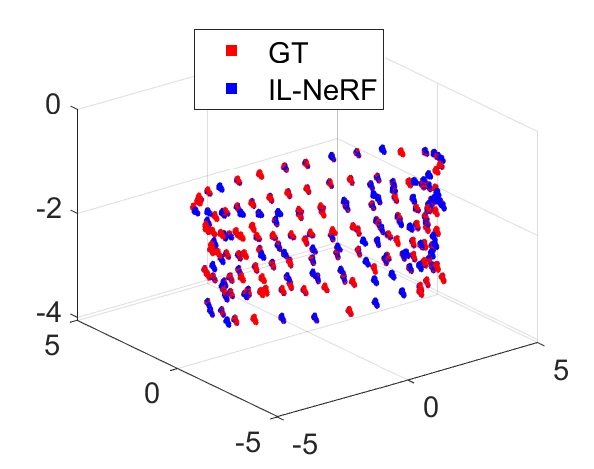}
    \caption{Garden}
    \end{subfigure}
    \begin{subfigure}{0.23\linewidth}
    \includegraphics[width=\linewidth]{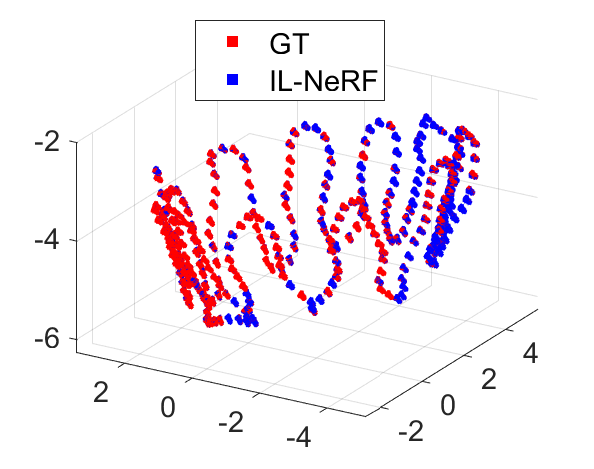}
    \caption{Bonsai}
    \end{subfigure}
    \begin{subfigure}{0.23\linewidth}
    \includegraphics[width=\linewidth]{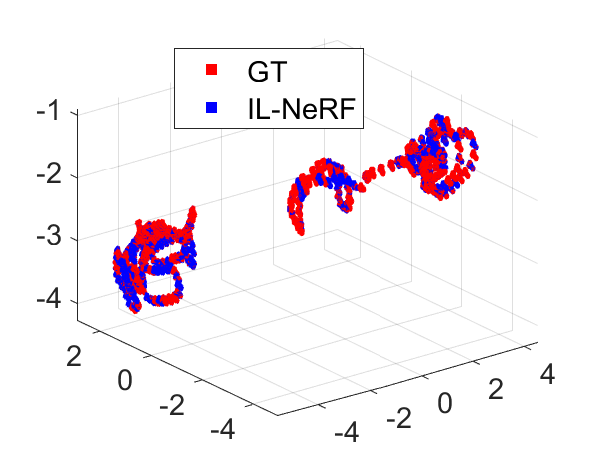}
    \caption{Room}
    \end{subfigure}
    \begin{subfigure}{0.23\linewidth}
    \includegraphics[width=\linewidth]{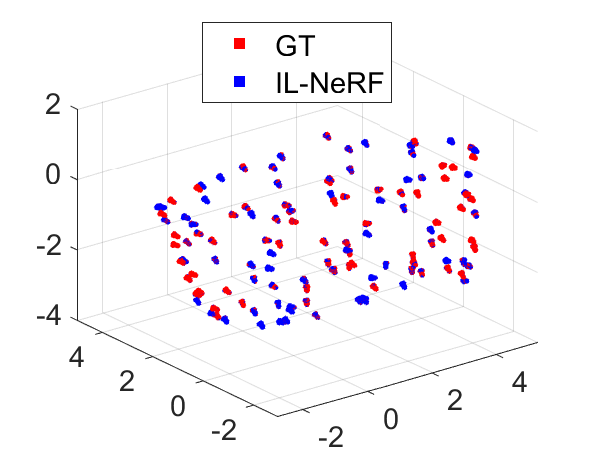}
    \caption{Stump}
    \end{subfigure}
    \begin{subfigure}{0.23\linewidth}
    \includegraphics[width=\linewidth]{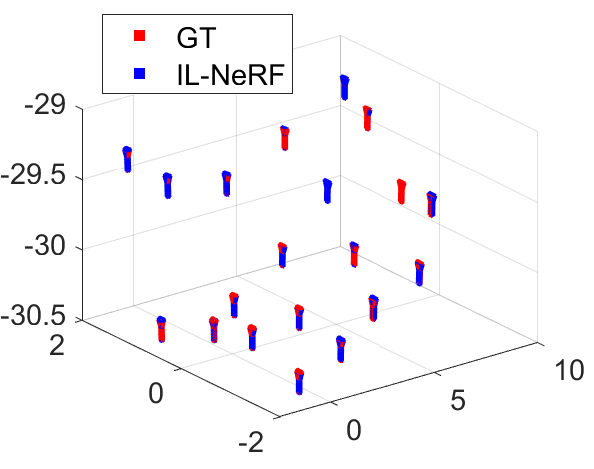}
    \caption{Fern}
    \end{subfigure}
    \begin{subfigure}{0.23\linewidth}
    \includegraphics[width=\linewidth]{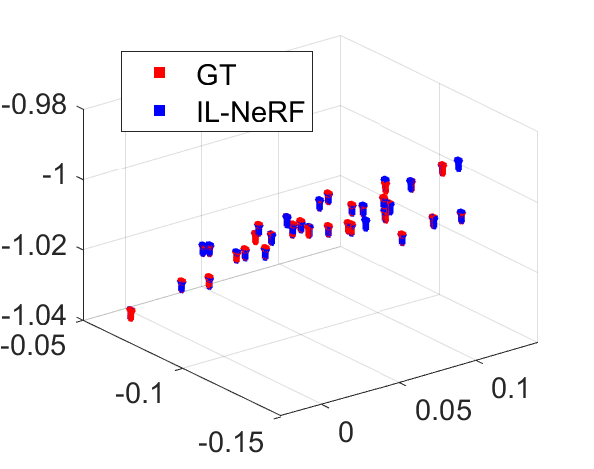}
    \caption{Flower}
    \end{subfigure}
    \begin{subfigure}{0.23\linewidth}
    \includegraphics[width=\linewidth]{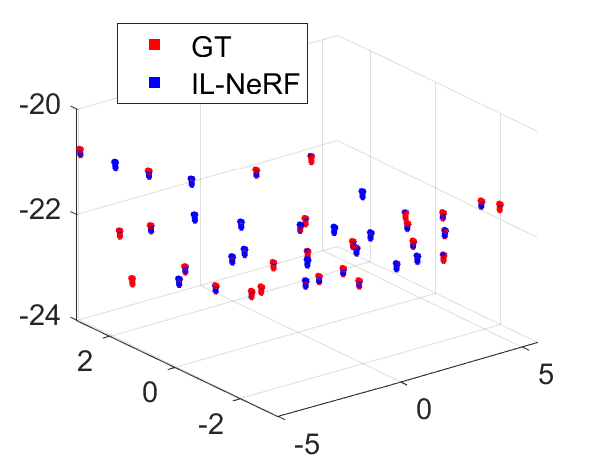}
    \caption{Fortress}
    \end{subfigure}
    \begin{subfigure}{0.23\linewidth}
    \includegraphics[width=\linewidth]{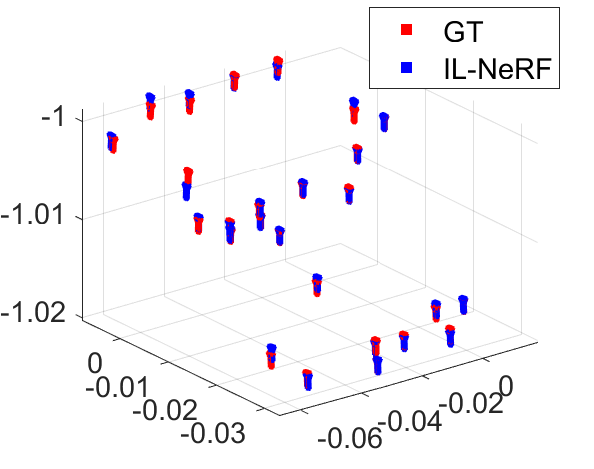}
    \caption{Leaves}
    \end{subfigure}
    \begin{subfigure}{0.23\linewidth}
    \includegraphics[width=\linewidth]{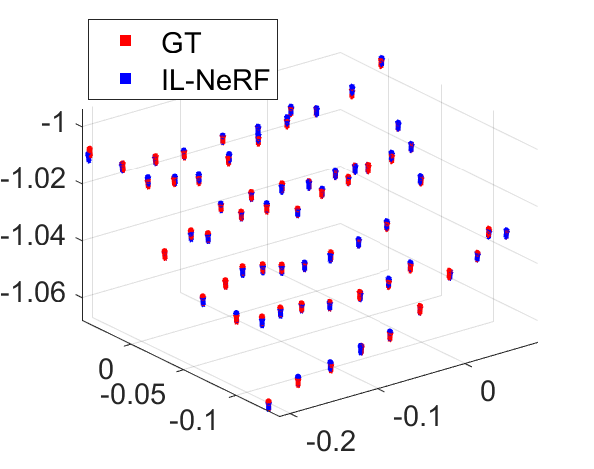}
    \caption{Horns}
    \end{subfigure}
    \begin{subfigure}{0.23\linewidth}
    \includegraphics[width=\linewidth]{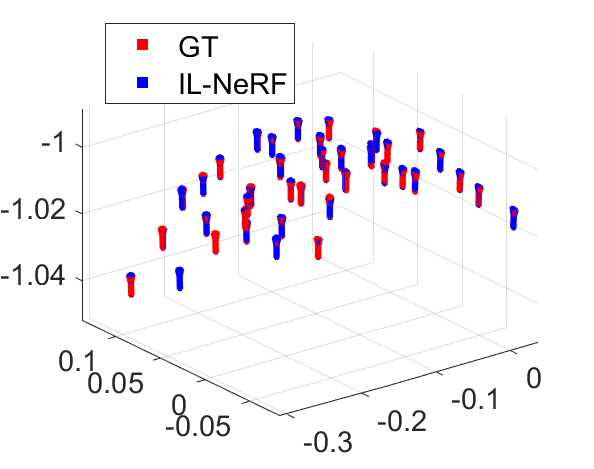}
    \caption{Meeting room}
    \end{subfigure}
    \begin{subfigure}{0.23\linewidth}
    \includegraphics[width=\linewidth]{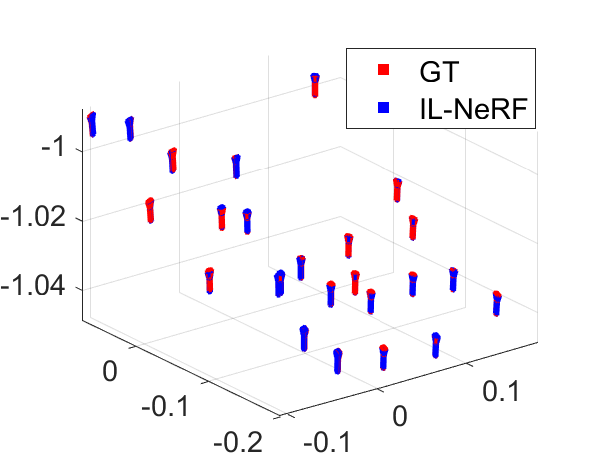}
    \caption{Orchids}
    \end{subfigure}
    \begin{subfigure}{0.23\linewidth}
    \includegraphics[width=\linewidth]{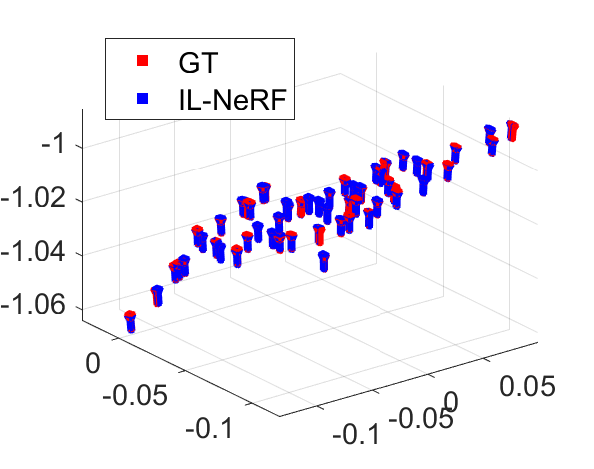}
    \caption{Trex}
    \end{subfigure}
    \begin{subfigure}{0.23\linewidth}
    \includegraphics[width=\linewidth]{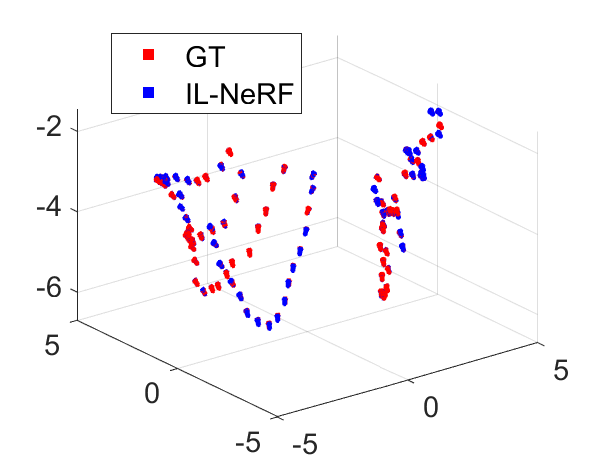}
    \caption{Pinecone}
    \end{subfigure}
    \caption{Camera pose estimation comparison. GT means the camera poses estimated by COLMAP from all the training images. IL-NeRF recovers accurate camera poses due to the help of incremental camera pose alignment.}
    \label{fig:supplement_cameraPose}
\end{figure*}

Figure \ref{fig:supplement_cameraPose} shows the camera pose trajectories of GT and IL-NeRF. We treat the COLMAP estimation from all training images as ground-truth (GT) camera poses. As we can see, IL-NeRF recovers accurate camera poses due to the help of incremental camera pose alignment.

\begin{table*}[!ht]
\centering
	\caption{Comparison of IL-
NeRF w/o TM, IL-NeRF w/o
PR and IL-NeRF. IL-NeRF
outperforms these two cases. }
    \begin{tabular}{cc|cccc}
    \toprule[1.5pt]
        Scene & Method & \multicolumn{4}{c}{\textbf{PSNR $\Uparrow$ / SSIM $\Uparrow$ / LPIPS $\Downarrow$}}\\
        \cline{3-6}
        & & $G^0$ & $G^1$ & $G^2$ & $G^3$\\
        \hline
        \multirow{4}{*}{Bicycle}& w/o TM & 22.90 / 0.62 / 0.33 & 13.64 / 0.32 / 0.66 & 11.86	/ 0.29 / 0.79 &	11.14 / 0.26 / 0.83 \\
        & w/o PR & 22.76 / 0.61 / 0.33 & 18.67 / 0.41 / 0.52 & 20.74 / 0.46 / 0.50 & 21.06 / 0.46 / 0.52 \\
        & IL-NeRF & \textbf{22.90} / \textbf{0.62} / \textbf{0.33}
        & \textbf{19.84} / \textbf{0.48} / \textbf{0.44} & \textbf{22.05} / \textbf{0.54} / \textbf{0.40} & \textbf{22.34} / \textbf{0.55} / \textbf{0.40} \\
        \hline
        \multirow{4}{*}{Bonsai}&  w/o TM & 33.54 / 0.93 / 0.07 & 21.13 / 0.59 / 0.18 & 19.48 / 0.46 / 0.31 & 18.40 / 0.40 / 0.37 \\
        & w/o PR & 33.30 / 0.93 / 0.07 & 28.30 / 0.84 / 0.18 & 27.80 / 0.81 / 0.21 & 27.33 / 0.80 / 0.22 \\
        & IL-NeRF & \textbf{33.54} / \textbf{0.93} / \textbf{0.07}
        & \textbf{30.73} / \textbf{0.89} / \textbf{0.12} & \textbf{29.77} / \textbf{0.86} / \textbf{0.16} & \textbf{28.96} / \textbf{0.85} / \textbf{0.18} \\
        \hline
        \multirow{4}{*}{Counter}&  w/o TM & 32.13 / 0.91 / 0.07 & 23.91 / 0.58 / 0.18 & 19.02 / 0.45 / 0.29 &	13.87 / 0.39 / 0.35\\
        & w/o PR & 32.12 / 0.91 / 0.07 & 27.89 / 0.83 / 0.13 & 27.05 / 0.81 / 0.16 & 26.47 / 0.79 / 0.17 \\
        & IL-NeRF & \textbf{32.13} / \textbf{0.91} / \textbf{0.07}
        & \textbf{29.63} / \textbf{0.87} / \textbf{0.12} & \textbf{28.56} / \textbf{0.85} / \textbf{0.15} & \textbf{27.82} / \textbf{0.83} / \textbf{0.17} \\ 
        \hline
        \multirow{4}{*}{Garden}& w/o TM & 24.73 / 0.73 / 0.19 & 17.05 / 0.46 / 0.33 & 13.37 / 0.37 / 0.45 &	15.76 / 0.31 / 0.47 \\
        & w/o PR & 24.70 / 0.71 / 0.20 & 23.34 / 0.67 / 0.20 & 23.17 / 0.69 / 0.21 & 22.42 / 0.67 / 0.23 \\
        & IL-NeRF & \textbf{24.73} / \textbf{0.73} / \textbf{0.19}
        & \textbf{24.80} / \textbf{0.70} / \textbf{0.22} & \textbf{24.86} / \textbf{0.69} / \textbf{0.23} & \textbf{24.82} / \textbf{0.67} / \textbf{0.23} \\
        \hline
        \multirow{4}{*}{Kitchen}& w/o TM & 31.27 / 0.92 / 0.07 & 21.08 / 0.59 / 0.15 & 16.05	/ 0.46 / 0.23 & 14.63 / 0.40 / 0.26\\
        & w/o PR & 31.17 / 0.91 / 0.08 & 28.54 / 0.84 / 0.12 & 27.86 / 0.82 / 0.15 & 27.48 / 0.78 / 0.18 \\
        & IL-NeRF & \textbf{31.27} / \textbf{0.92} / \textbf{0.07}
        & \textbf{30.66} / \textbf{0.89} / \textbf{0.10} & \textbf{29.84} / \textbf{0.87} / \textbf{0.12} & \textbf{29.34} / \textbf{0.86} / \textbf{0.13} \\
        \hline
        \multirow{4}{*}{Room}& w/o TM & 36.04 / 0.96 / 0.04 & 	27.45 / 0.62 / 0.06 &	17.40 / 0.49 /	0.13 & 	19.98 /	0.43 / 0.18\\
        & w/o PR & 35.98 / 0.96 / 0.04 & 32.67 / 0.92 / 0.07 & 31.12 / 0.86 / 0.17 & 30.35 / 0.84 / 0.13 \\
        & IL-NeRF & \textbf{36.04} / \textbf{0.96} / \textbf{0.04}
        & \textbf{34.02} / \textbf{0.94} / \textbf{0.04} & \textbf{32.35} / \textbf{0.92} / \textbf{0.07} & \textbf{31.45} / \textbf{0.91} / \textbf{0.09} \\
        \hline
        \multirow{4}{*}{Stump}& w/o TM & 25.96 /	0.77 /	0.28 &	20.78 / 0.44 / 0.48 &	16.71 / 0.32 / 0.73	& 15.81 /	0.27 / 0.80\\
        & w/o PR & 25.62 / 0.77 / 0.28 & 24.25 / 0.58 / 0.37 & 23.77 / 0.53 / 0.43 & 22.43 / 0.50 / 0.46 \\
        & IL-NeRF & \textbf{25.96} / \textbf{0.77} / \textbf{0.28}
        & \textbf{25.75} / \textbf{0.66} / \textbf{0.32} & \textbf{25.09} / \textbf{0.60} / \textbf{0.37} & \textbf{24.89} / \textbf{0.58} / \textbf{0.39} \\
        \toprule[1.5pt]
    \end{tabular}
\label{table:supplement_accuracyOnMipNeRF360}
\end{table*}

\begin{table*}[!ht]
\centering
	\caption{Comparison of IL-
NeRF w/o TM, IL-NeRF w/o
PR and IL-NeRF. IL-NeRF
outperforms these two cases. }
    \begin{tabular}{cc|cccc}
    \toprule[1.5pt]
        Scene & Method & \multicolumn{4}{c}{\textbf{PSNR $\Uparrow$ / SSIM $\Uparrow$ / LPIPS $\Downarrow$}}\\
        \cline{3-6}
        & & $G^0$ & $G^1$ & $G^2$ & $G^3$\\
        \hline
        \multirow{4}{*}{Fern}& w/o TM & 29.30 / 0.90 / 0.06	& 18.34 / 0.58 / 0.15 & 13.79 / 0.43 / 0.25 & 	16.04 /	0.37 /	0.31\\
        & w/o PR & 29.19 / 0.90 / 0.06 &  25.37 / 0.82 / 0.11 & 24.62 / 0.77 / 0.19 & 24.77 / 0.78 / 0.20\\
        & IL-NeRF &  \textbf{29.30} / \textbf{0.90} / \textbf{0.06} & \textbf{26.68} / \textbf{0.87} / \textbf{0.10} & \textbf{25.63} / \textbf{0.81} / \textbf{0.13} & \textbf{25.26} / \textbf{0.80} / \textbf{0.15} \\
        \hline
        \multirow{4}{*}{Flower}& w/o TM & 34.22 / 0.96 / 0.01	& 25.67 / 0.62 /	0.15 &	20.72 / 0.50 / 0.39	& 14.37 / 0.43 /	0.44\\
        & w/o PR &  34.12 / 0.96 / 0.01 & 30.82 / 0.92 / 0.02 & 30.69 / 0.91 / 0.03 & 28.41 / 0.89 / 0.03\\
        & IL-NeRF & \textbf{34.22} / \textbf{0.96} / \textbf{0.01} & \textbf{31.81} / \textbf{0.94} / \textbf{0.01} & \textbf{31.11} / \textbf{0.94} / \textbf{0.02} & \textbf{30.49} / \textbf{0.93} / \textbf{0.02}  \\
        \hline
        \multirow{4}{*}{Fortress}& w/o TM & 31.6 / 0.85 / 0.11 &	21.33 / 0.56 /	0.15 &	20.20 /	0.45 / 0.21	& 16.71 /	0.39 /	0.34\\
        & w/o PR & 31.56 / 0.85 / 0.11 & 30.62 / 0.82 / 0.14 & 29.75 / 0.79 / 0.15 & 28.78 / 0.76 / 0.17\\
        & IL-NeRF & \textbf{31.69} / \textbf{0.85} / \textbf{0.11} & \textbf{31.02} / \textbf{0.84} / \textbf{0.10} & \textbf{30.33} / \textbf{0.84} / \textbf{0.11} & \textbf{29.45} / \textbf{0.83} / \textbf{0.12}\\ 
        \hline
        \multirow{4}{*}{Horns}& w/o TM & 31.6 / 0.85 / 0.11 &	21.33 / 0.56 /	0.15 &	20.20 /	0.45 / 0.21	& 16.71 /	0.39 /	0.34\\
        & w/o PR & 29.92 /	0.89 / 0.07 &	20.28 / 0.59 / 0.20 &	15.61 / 0.47 /	0.33 &	14.44 /	0.41 / 0.48\\
        & IL-NeRF & \textbf{29.92} / \textbf{0.89} / \textbf{0.07} & \textbf{29.50} / \textbf{0.89} / \textbf{0.07} & \textbf{29.01} / \textbf{0.89} / \textbf{0.07} & \textbf{28.96} / \textbf{0.87} / \textbf{0.09}\\
        \hline
        \multirow{4}{*}{Leaves}& w/o TM & 25.62 / 0.90	/ 0.06 & 17.01 / 0.58 /	0.20 &	13.05 / 0.46 /	0.33 &	11.17 /	0.40 /	0.46\\
        & w/o PR &  25.51 / 0.90 / 0.06 & 24.20 / 0.87 / 0.06 & 23.77 / 0.84 / 0.08 & 22.69 / 0.83 / 0.09\\
        & IL-NeRF & \textbf{25.62} / \textbf{0.90} / \textbf{0.06} & \textbf{24.74} / \textbf{0.88} / \textbf{0.07} & \textbf{24.26} / \textbf{0.87} / \textbf{0.07} & \textbf{23.88} / \textbf{0.86} / \textbf{0.08}\\
        \hline
        \multirow{4}{*}{Orchids}& w/o TM & 25.78 / 0.86 / 0.08 &	19.50 / 0.54	/ 0.15 & 15.85 / 0.42 /	0.23 &	12.03 /	0.36 /	0.46\\
        & w/o PR &  25.68 / 0.85 / 0.08 & 23.08 / 0.79/ 0.10 & 22.96 / 0.75 / 0.12 & 22.23 / 0.72 / 0.13 \\
        & IL-NeRF & \textbf{25.78} / \textbf{0.86} / \textbf{0.08} & \textbf{24.17} / \textbf{0.82} / \textbf{0.10} & \textbf{23.89} / \textbf{0.80} / \textbf{0.12} & \textbf{23.67} / \textbf{0.77} / \textbf{0.13} \\
        \hline
        \multirow{4}{*}{Room}& w/o TM & 32.50 /	0.92 /	0.08 &	25.63 / 0.61 / 0.12 &	20.99 /	0.49 / 0.25	& 16.25 /	0.43 /	0.34\\
        & w/o PR & 31.14 / 0.92 / 0.09 & 30.37 / 0.90 / 0.09 & 30.56 / 0.91 / 0.09 & 30.58 / 0.90 / 0.08 \\
        & IL-NeRF & \textbf{32.50} / \textbf{0.92} / \textbf{0.08} & \textbf{31.76} / \textbf{0.92} / \textbf{0.08} &\textbf{31.58} / \textbf{0.92} / \textbf{0.08} & \textbf{31.88} / \textbf{0.92} / \textbf{0.07}\\
        \hline
        \multirow{4}{*}{Trex}& w/o TM & 28.70 /	0.90 / 	0.07 &	19.35 /	0.60 / 0.19 & 	15.01 /	0.48 / 0.33 &	13.86 /	0.42 / 0.42\\
        & w/o PR &  28.50 / 0.90 / 0.07 & 27.01 / 0.87 / 0.09 & 26.26 / 0.88 / 0.07 & 26.54 / 0.88 / 0.08 \\ 
        & IL-NeRF & \textbf{28.70} / \textbf{0.90} / \textbf{0.07} & \textbf{28.14} / \textbf{0.90} / \textbf{0.06} & \textbf{27.90} / \textbf{0.90} / \textbf{0.07} & \textbf{27.81} / \textbf{0.90} / \textbf{0.06}\\
        \toprule[1.5pt]
    \end{tabular}
\label{table:supplement_accuracyOnLLFF}
\end{table*}

\begin{table*}[!ht]
\centering
	\caption{Comparison of IL-
NeRF w/o TM, IL-NeRF w/o
PR and IL-NeRF. IL-NeRF
outperforms these two cases. }
    \begin{tabular}{cc|cccc}
    \toprule[1.5pt]
        Scene & Method & \multicolumn{4}{c}{\textbf{PSNR $\Uparrow$ / SSIM $\Uparrow$ / LPIPS $\Downarrow$}}\\
        \cline{3-6}
        & & $G^0$ & $G^1$ & $G^2$ & $G^3$\\
        \hline
        \multirow{4}{*}{Pinecone}& w/o TM & 26.31 / 0.87 / 0.10 & 19.82 / 0.52 / 0.25 & 15.84 / 0.39 / 0.39 & 	14.56 /	0.34 / 0.47\\
        & w/o PR & 26.22 / 0.84 / 0.16 & 23.87 / 0.61 / 0.22 & 22.24 / 0.66 / 0.28 & 21.13 / 0.66 / 0.32 \\
        & IL-NeRF & \textbf{26.31} / \textbf{0.87} / \textbf{0.10}
        & \textbf{24.56} / \textbf{0.78} / \textbf{0.17} & \textbf{23.78} / \textbf{0.74} / \textbf{0.20} & \textbf{22.93} / \textbf{0.72} / \textbf{0.23} \\
        \hline
        \multirow{4}{*}{Vasedeck}& w/o TM & 29.48 / 0.86 / 0.07 & 22.09 / 0.54 / 0.25 & 16.05 / 0.40 / 0.35 & 13.04 / 0.35 / 0.37\\
        & w/o PR & 29.03 / 0.85 / 0.07 & 25.30 / 0.74 / 0.18 & 24.80 / 0.68 / 0.21 & 24.33 / 0.63 / 0.23 \\
        & IL-NeRF & \textbf{29.48} / \textbf{0.86} / \textbf{0.07}
        & \textbf{27.38} / \textbf{0.82} / \textbf{0.10} & \textbf{26.11} / \textbf{0.76} / \textbf{0.14} & \textbf{26.15} / \textbf{0.75} / \textbf{0.17} \\
        \toprule[1.5pt]
    \end{tabular}
\label{table:supplement_accuracyOnNeRFReal360}
\end{table*}

\section{Limitation}
For large-scale scenes with limited overlap between views in the training dataset, the performance of IL-NeRF may be suboptimal because the limited overlap between views can result in significant errors or even the inability to calculate the transfer matrices during the camera coordinate alignment.

\end{document}